\documentclass[]{fairmeta}

\usepackage[utf8]{inputenc}

\usepackage[ruled, vlined, linesnumbered]{algorithm2e}
\usepackage{bm}
\usepackage{bbm}
\usepackage{amsthm}
\usepackage{amsmath}
\usepackage{mathrsfs} 
\usepackage{tabularx}
\usepackage{graphicx}
\usepackage{multirow}
\usepackage{xspace}
\usepackage{enumitem}
\usepackage{placeins}
\usepackage{soul}
\usepackage{mathtools}
\usepackage{booktabs}
\usepackage{hyperref}
\usepackage{subcaption}
\usepackage{pgfplots}
\usepackage{csquotes}
\usepackage{cleveref}
\usepackage{balance}
\usepackage{multicol}
\usepackage{makecell}
\usepackage{amssymb}
\usepackage[switch]{lineno}
\DeclareUnicodeCharacter{2212}{−}
\usepgfplotslibrary{groupplots,dateplot}
\usetikzlibrary{patterns,shapes.arrows}
\pgfplotsset{compat=newest}

\usepackage{fvextra}
\usepackage[most]{tcolorbox}
\newtcolorbox{prompt}[1]{colback=gray!20,colframe=gray!50!black,fonttitle=\bfseries,title=#1}

\newcommand{\ours}{{Dr.\,Zero}\xspace}
\newcommand{\hrpo}{{HRPO}\xspace}
\newcommand{\hrpolong}{{hop-grouped relative policy optimization}\xspace}

\usepackage{fontawesome5} 
\usepackage{xcolor}

\definecolor{chatassistant}{HTML}{E6F3FF}
\definecolor{chatuser}{HTML}{F0F0F0}

\newtcolorbox{AssistantBox}{
  enhanced,
  colback=chatassistant,
  colframe=blue!40!black,
  fontupper=\footnotesize\ttfamily, 
  fonttitle=\bfseries\scriptsize,
  boxrule=0.4pt,
  left=2pt, right=2pt, top=2pt, bottom=2pt,
  boxsep=1pt,
  arc=1mm,
  title={\quad Assistant}
}

\newtcolorbox{UserBox}{
  enhanced,
  colback=chatuser,
  colframe=gray!60!black,
  fontupper=\footnotesize\ttfamily, 
  fonttitle=\bfseries\scriptsize,
  boxrule=0.4pt,
  left=2pt, right=2pt, top=2pt, bottom=2pt,
  boxsep=1pt,
  arc=1mm,
  title={\quad User / Tool}
}

\newtcolorbox{UserBoxS}{
  enhanced,
  colback=chatuser,
  colframe=gray!60!black,
  fontupper=\scriptsize\ttfamily, 
  fonttitle=\bfseries\scriptsize,
  boxrule=0.4pt,
  left=2pt, right=2pt, top=2pt, bottom=2pt,
  boxsep=1pt,
  arc=1mm,
  title={\quad User / Tool}
}

\newtcolorbox{SystemBox}{
  enhanced,
  colback=chatuser,
  colframe=gray!60!black,
  fontupper=\footnotesize\ttfamily, 
  fonttitle=\bfseries\scriptsize,
  boxrule=0.4pt,
  left=2pt, right=2pt, top=2pt, bottom=2pt,
  boxsep=1pt,
  arc=1mm,
  title={\quad System}
}

\newtcolorbox{SystemBoxS}{
  enhanced,
  colback=chatuser,
  colframe=gray!60!black,
  fontupper=\scriptsize\ttfamily, 
  fonttitle=\bfseries\scriptsize,
  boxrule=0.4pt,
  left=2pt, right=2pt, top=2pt, bottom=2pt,
  boxsep=1pt,
  arc=1mm,
  title={\quad System}
}

\DefineVerbatimEnvironment{ChatVerb}{Verbatim}{
  breaklines=true,
  breakanywhere=true,
  fontsize=\footnotesize
}

\DefineVerbatimEnvironment{ChatVerbS}{Verbatim}{
  breaklines=true,
  breakanywhere=true,
  fontsize=\scriptsize
}

\title{\ours: Self-Evolving Search Agents without Training Data}

\author[1,2*]{Zhenrui Yue}
\author[1]{Kartikeya Upasani}
\author[1]{Xianjun Yang}
\author[2]{Suyu Ge}
\author[1]{Shaoliang Nie}
\author[1]{Yuning Mao}
\author[1]{Zhe Liu}
\author[2]{Dong Wang}

\affiliation[1]{Meta Superintelligence Labs}
\affiliation[2]{University of Illinois Urbana-Champaign}

\abstract{
As high-quality data becomes increasingly difficult to obtain, self-evolution without curated training data has emerged as a promising paradigm. This approach allows large language models (LLMs) to autonomously generate and solve complex problems, thereby improving their reasoning capabilities. However, multi-turn search agents struggle in this setting due to limited question diversity and the substantial compute required for multi-step reasoning and tool use. In this work, we introduce \ours, a framework that enables search agents to effectively self-evolve without human-annotated training data, relying solely on an external search engine as their knowledge environment. In particular, we design a self-evolution feedback loop where a proposer generates structurally diverse questions to train a solver initialized from the same base model. As the solver evolves, it incentivizes the proposer to produce increasingly difficult yet solvable tasks, thus establishing an automated curriculum to refine both agents. To enhance training efficiency, we also introduce \hrpolong (\hrpo). This method clusters structurally similar questions to construct group-level baselines, effectively minimizing the sampling overhead in evaluating each query's individual difficulty and solvability. Consequently, \hrpo significantly reduces the compute requirements for proposer training and reward estimation without compromising performance or stability. Extensive experimental results demonstrate that \ours matches or surpasses fully supervised search agents on several question answering benchmarks, showing that strong agentic search and evidence-grounded reasoning can emerge solely through self-evolution.
}

\date{\today}
\metadata[Code]{\url{https://github.com/facebookresearch/drzero}}
\correspondence{\email{\{zhenrui3, dwang24\}@illinois.edu, kart@meta.com}}
\contribution{*Work done while at Meta}

\begin{document}

\maketitle

\section{Introduction}

\begin{figure}[t]
    \centering
    \includegraphics[trim=7cm 4cm 7cm 4cm, clip, width=0.55\columnwidth]{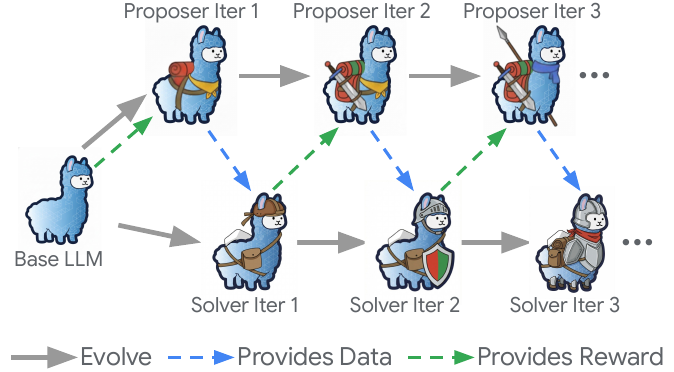}
    \caption{The self-evolving LLM training framework~\citep{huang2025r} that iteratively trains a proposer model and a solver model with minimal supervision.}
    \label{fig:intro}
\end{figure}

Self-evolving language agents have emerged as a promising paradigm to autonomously enhance model performance by iteratively proposing problems, bootstrapping solutions, and learning from prior experiences~\citep{liu2025spiral, zhang2025evolvesearch}. However, existing methods typically rely on extensive curated questions~(prompts) to drive exploration, maintaining the need for the costly data curation process~\citep{chen2024self, wang2025socratic}. To overcome this bottleneck, an automated proposer can be employed to generate synthetic training data, thereby making self-evolution possible with minimal or even zero human input~\citep{zhao2025absolute, chen2025self}. As shown in \Cref{fig:intro}, \citet{huang2025r} design a proposer-solver co-evolution framework to iteratively bootstrap questions and rationales, thereby achieving meaningful performance gains without access to any human-annotated datasets.

Nevertheless, self-evolving language agents primarily focus on mathematical or specific reasoning tasks~\citep{zhao2025absolute, huang2025r, wang2025socratic}. In these constrained domains, questions are often narrowly defined, allowing agents to achieve considerable gains even with limited or low-diversity training data~\citep{zuo2025ttrl, wang2025reinforcement}. Yet for open-domain questions, self-evolving search agents remain substantially under-explored. While some approaches exist, they typically rely on human questions, extensive contexts or ground truth annotations~\citep{jiang2025s3, sun2025zerosearch, lu2025search}. To address this limitation, we focus on self-evolution without curated QA training data for learning search agents. Distinct from prior work, we eliminate the need for human-written training questions or answer annotations, while using an external search engine and its indexed corpus as the knowledge environment for the agent's self-evolution.

We analyze existing data-free frameworks~\citep{huang2025r, chen2025self} and identify two primary limitations in proposer training:~(1)~the trained proposer suffers from limited diversity, exhibiting a bias toward simple, one-hop questions (see \Cref{sec:question-analysis}); and~(2)~their proposer objectives do not explicitly adapt query difficulty as the solver's capabilities advance. As a result, such approaches yield moderate performance gains on trivial one-hop tasks but struggle to match supervised baselines on complex multi-hop queries~(see \Cref{sec:exp}). Moreover, the standard group relative policy optimization~(GRPO)~\citep{shao2024deepseekmath} significantly increases training compute in self-evolution as it requires nested sampling: generating multiple queries and subsequently producing multiple responses for each question. For instance, ~\citet{chen2025self} propose a unified model that acts as both proposer and solver for simple math and coding tasks. However, its reliance on nested sampling renders this approach computationally prohibitive for search agents and creates bottlenecks during complex reasoning and search interactions.

In this paper, we investigate self-evolving search agents without curated QA training data. We propose \ul{D}eep\ul{R}esearch-\ul{Zero}~(\ours), a framework that leverages external search engines to improve both proposer and solver performance without human-written training questions or answers. Specifically, we introduce a multi-turn tool-use rollout pipeline that enables the trained proposer to improve question generation and produce complex, multi-hop questions. Furthermore, we design a difficulty-guided reward that incentivizes the proposer to utilize the search engine and generate challenging yet verifiable queries. As computing the reward involves sampling multiple reasoning trajectories from the solver, we also propose \hrpolong~(\hrpo), which clusters structurally similar questions to provide a more robust baseline for advantage estimation. Thus, \ours avoids nested sampling by generating one question per seed prompt and estimating its baseline across questions with the same hop structure, rather than sampling multiple question candidates from each seed prompt as in standard GRPO. Combined with further training and inference optimizations, \ours achieves substantial gains compared to the base LLM. Additionally, through multiple training iterations, the model matches or even outperforms supervised RL baselines, showing that complex reasoning and search capabilities can be induced without human-annotated training questions in the evaluated short-form, verifiable QA setting. We summarize our contributions as follows:
\begin{enumerate}
    \item We propose \ul{D}eep\ul{R}esearch-\ul{Zero}~(\ours), a unified framework that leverages external search engines to autonomously improve both proposer and solver performance. Our approach integrates a refined multi-turn tool-use pipeline with a difficulty-guided reward to produce complex, multi-hop questions. 
    \item We introduce \hrpolong~(\hrpo), a novel optimization method that clusters structurally similar questions to provide a robust group-level baseline for advantage estimation. This optimization ensures effective training while eliminating the need for expensive nested sampling in self-evolution. 
    \item We empirically demonstrate the effectiveness of \ours through extensive experiments, showing that the proposed self-evolution paradigm yields significant performance gains. Without curated QA training data, our search agents match the aggregate performance of a fully supervised search agent at the 3B scale and outperform supervised baselines on several individual benchmarks.
\end{enumerate}
\section{Related Work}

\subsection{Reinforcement Learning}
Reinforcement learning (RL) improves agent performance by learning from previous experience and maximizing cumulative rewards~\citep{sutton1998reinforcement}. In the context of LLMs, RL is frequently implemented using policy gradient algorithms~\citep{sutton1999policy, ouyang2022training}. For example, actor-critic methods such as PPO employ a learned critic to estimate a value baseline, thereby reducing gradient variance~\citep{mnih2016asynchronous, schulman2017proximal}. A simpler offline alternative is direct preference optimization (DPO)~\citep{rafailov2023direct}, which directly optimizes language models on pairwise preference data. Recently, group-based methods like GRPO have been adopted for their ability to construct low-variance baselines from multiple responses~\citep{ahmadian2024back, shao2024deepseekmath, hu2025reinforce++}. Building upon these algorithmic improvements, LLMs have demonstrated significant potential across diverse model architectures as well as reasoning-intensive tasks~\citep{yu2025dapo, huang2025vision, zheng2025group}.

\subsection{Search-Augmented LLMs}
Search and retrieval augmentation can improve language modeling by integrating external knowledge~\citep{lewis2020retrieval, guu2020retrieval}. Furthermore, iteratively retrieving relevant documents can substantially enhance LLM performance on complex questions~\citep{yoran2023making, yue2024inference, wang2025archrag}. A notable example is IRCoT, where \citet{trivedi2023interleaving} exploit multi-step retrieval to optimize answer accuracy on knowledge-intensive tasks. In agentic LLMs, search engines can be incorporated as optional tools to enable adaptive retrieval and reasoning, leading to enhanced multi-hop reasoning performance~\citep{jin2025search, zheng2025deepresearcher, zhang2025web}. Nevertheless, existing methods primarily focus on supervised, verifiable settings and rely on extensive human queries and annotations for training. Concurrent with our work, \citet{lu2025search} propose using self-play to improve search-augmented LLMs. However, their method still depends on ground truth labels and numerous human-written examples. In contrast, we use no curated QA training data and rely on external search as the knowledge environment for self-improvement.

\subsection{Self-Evolving LLMs}
Self-evolving LLMs autonomously enhance model capabilities by iteratively generating and learning from their own experiences~\citep{pang2024iterative, liu2025spiral}. Early approaches utilize self-play mechanisms where the model acts as both the generator and the evaluator to refine its policy without human annotations~\citep{openai2021asymmetric, chen2024self, wu2024self}. For instance, self-rewarding LLMs~\citep{yuan2024self} employ iterative training loops where the model judges its own outputs to construct preference data for optimization. To further eliminate the reliance on external prompts, recent works have introduced data-free self-evolution frameworks~\citep{zhao2025absolute, chen2025self, liu2025spice, xia2025agent0}. Notably, R-Zero~\citep{huang2025r} adopts a proposer-solver design to co-evolve their performance from scratch, effectively creating a self-improving curriculum. Although promising, existing methods are computationally intensive for multi-turn search agents and often underperform on open-domain questions due to the lack of diverse data and tool integration. To bridge this gap, we introduce \ours, an efficient self-evolution framework without curated QA training data that enables search agents to match or even exceed the performance of strong supervised baselines.

\section{Methodology}

\subsection{Setup}
\begin{figure*}
    \centering
    \includegraphics[trim=3cm 3.4cm 2.4cm 3.4cm, clip, width=1.0\textwidth]{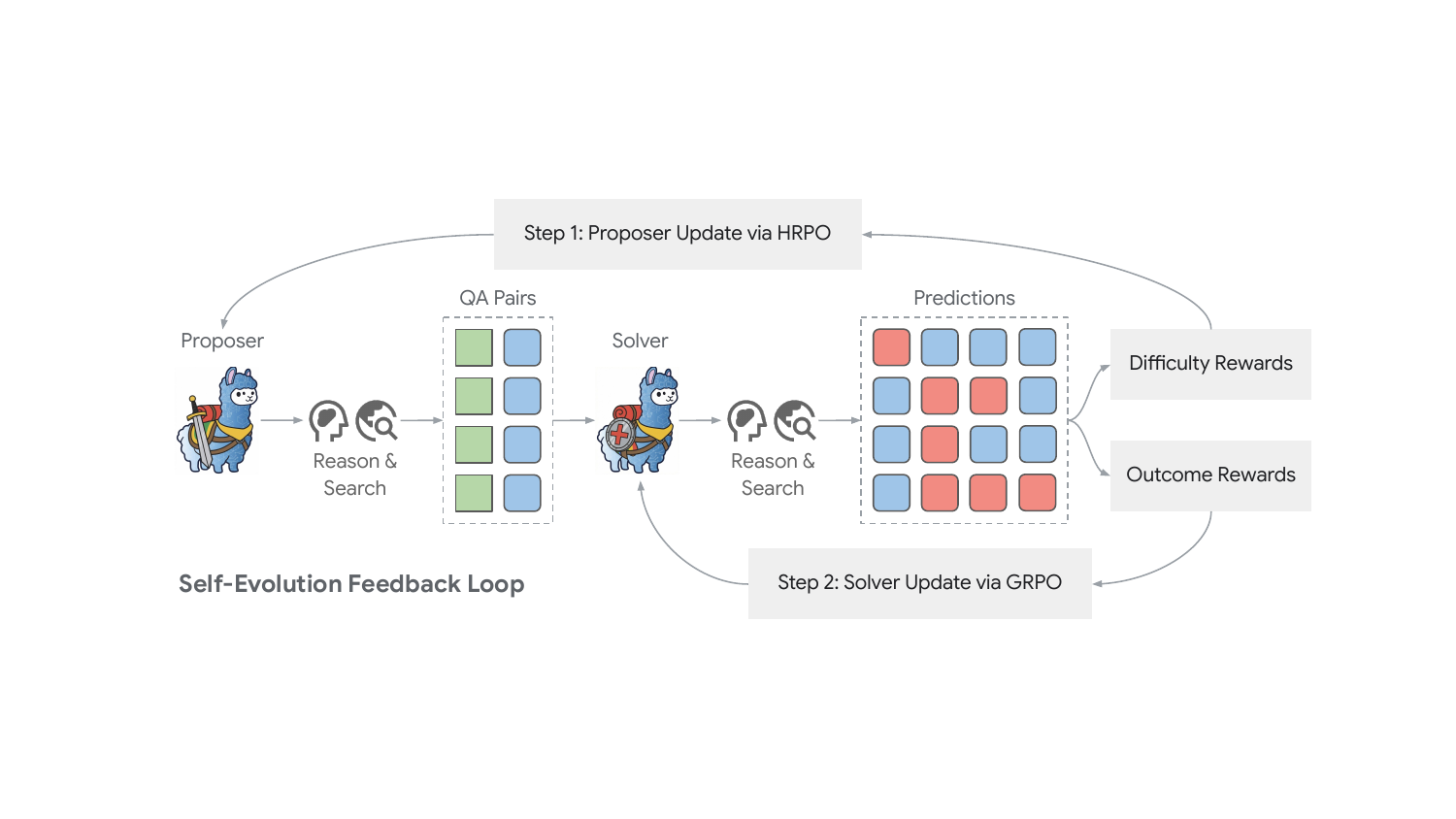}
    \caption{The \ours self-evolution feedback loop. Guided by solver feedback, the proposer synthesizes verifiable and challenging queries, continuously enhancing the solver's search and reasoning capabilities.}
    \label{fig:method}
\end{figure*}

We employ a proposer-solver self-evolution framework where both models function as search agents capable of leveraging external knowledge. Equipped with the external search engine $\mathcal{R}$, the proposer $\pi_\theta$ and the solver $\pi_\phi$ are trained to maximize their respective expected rewards:
\begin{equation}
\begin{aligned}
\mathrm{Proposer:} \quad & \mathbb{E}_{\substack{(x, y) \sim \pi_\theta(\cdot | \mathcal{R}), \{ \hat{y}_i \}_{i=1}^n} \sim \pi_\phi(\cdot | x, \mathcal{R})} [ r (y, \{ \hat{y}_i \}_{i=1}^n) ], \\
\mathrm{Solver:} \quad &\mathbb{E}_{\substack{(x, y) \sim \pi_\theta(\cdot | \mathcal{R}), \hat{y} \sim \pi_\phi(\cdot | x, \mathcal{R})}} [ \mathbb{I}(y = \hat{y}) ],
\label{eq:objective}
\end{aligned}
\end{equation} 
where $r$ denotes the proposer reward and $\mathbb{I}$ is the indicator function. In contrast to the solver's simple outcome-based reward, the proposer reward is defined over the distribution of predicted answers (i.e., $\{ \hat{y}_i \}_{i=1}^n$). If all predictions are correct, the question is considered trivial, whereas if none are correct, the question is likely too difficult for the solver. To enable self-evolution of $\pi_\theta$ and $\pi_\phi$, we iteratively optimize both components in a symbiotic loop: the proposer learns to synthesize diverse and challenging questions, while the solver enhances its reasoning abilities by learning from these questions. The improved solver performance subsequently encourages the proposer to generate increasingly complex queries, forming a continuously evolving curriculum (see \Cref{fig:method}). We initialize both models from the same base LLM and rely exclusively on the search tool ($\mathcal{R}$) for external knowledge. To adhere to our setting without curated QA training data, we avoid utilizing any demonstrations, human-written questions or annotated answers in our framework.

\subsection{Proposer Training}
Existing self-evolution methods primarily target specialized domains (e.g., math, coding) to enhance LLM performance without external data~\citep{zhao2025absolute, huang2025r, chen2025self}. However, for open-domain question answering, we find that such methods tend to generate structurally homogeneous one-hop queries, which limits solver performance gains to simple reasoning questions. Even when equipped with a multi-turn search tool, the solver yields only marginal improvements on multi-hop queries (see \Cref{sec:exp}). Furthermore, since the proposer reward necessitates multiple solver predictions to assess query difficulty and solvability, optimizing the proposer with a standard RL algorithm like GRPO becomes highly inefficient~\citep{shao2024deepseekmath}. Generating $m$ candidate QA pairs from each seed prompt and evaluating each with $n$ solver predictions requires $m + mn = m(n+1)$ rollouts. In our implementation, this corresponds to 4 proposer rollouts and $4 \times 4$ solver rollouts (20 total). Combined with the high latency of multi-turn rollouts, this scaling bottleneck renders existing approaches impractical for self-evolving agents that require complex tool interactions.

Motivated by these limitations, we propose \hrpolong (\hrpo) to train the proposer model. Instead of sampling multiple responses for a single seed prompt, \hrpo calculates advantages by grouping structurally similar questions across prompts. Specifically, we cluster the generated QA pairs by their cross-hop complexity, denoted by the number of hops $h \in \mathcal{H}$. Questions with fewer hops are typically simpler, whereas higher-hop questions demand extensive search and multi-step reasoning. Here, hop count is used only as a structural grouping variable; empirical question difficulty is determined by the solver pass rate in \Cref{eq:reward}. This hop-specific normalization of returns produces low-variance advantage estimates while avoiding the computational cost of sampling multiple candidate questions per seed prompt. With five solver predictions, \hrpo requires one proposer rollout and five solver rollouts (6 total). \hrpo can be formulated with:
\begin{equation}
\begin{aligned}
    \mathcal{J}(\theta) &= \mathbb{E}_{\substack{\{ (x_i, y_i) \sim \pi_\theta(\cdot | \mathcal{R}), \{ \hat{y}_{i, k} \}_{k=1}^n \sim \pi_\phi(\cdot | x_i, \mathcal{R}) \}_{i=1}^N}} \\
    &[ \frac{1}{N} \sum_{h \in \mathcal{H}} \sum_{i \in \mathcal{I}_h} \log \pi_\theta( x_i, y_i | \mathcal{R}) A_{i,h} ] - \beta \mathbb{D}_{KL},
\label{eq:hrpo}
\end{aligned}
\end{equation}
where $N$ denotes the size of the sampled batch, and $\beta$ is the hyperparameter controlling the KL regularizer. $A_{i,h}$ denotes the advantage of the generated QA pair $(x_i, y_i)$, computed by standardizing the solver's reward scores over all $h$-hop questions:
\begin{equation}
    A_{i,h} = \frac{r_i - \mathbb{E}_{j \in \mathcal{I}_h}[r_j]}{\sqrt{\mathbb{V}\text{ar}_{j \in \mathcal{I}_h}[r_j]} + \delta}.
\label{eq:advantage}
\end{equation}
For optimal proposer performance and training efficiency, we adopt a strictly on-policy framework and omit ratio clipping. While single-response methods like REINFORCE++ reduce sampling costs, we find that a global baseline becomes unstable when processing diverse query structures. This mismatch induces high variance in policy gradients, frequently leading to training failures. In contrast, \hrpo mitigates this issue by computing relative advantages among structurally aligned trajectories.

To utilize solver signals for training the proposer $\pi_\theta$, we design a specialized reward function to encourage both verifiability (the task must be solvable) and difficulty (the task must not be trivial). We leverage the solver's pass rate on the generated questions as a proxy for these properties. Let $k$ denote the number of correct solutions out of $n$ sampled attempts; we penalize instances where the solver either fails completely ($k = 0$) or succeeds trivially ($k = n$), incentivizing the generation of questions that maximize:
\begin{equation}
    r (y, \{ \hat{y}_i \}_{i=1}^n) = \mathbb{I}(0 < k < n) \frac{n - k}{n - 1} + r^f, \; \text{with} \; k = \sum_{i=1}^n \mathbb{I}(y = \hat{y}_i),
\label{eq:reward}
\end{equation}
here, the reward is maximized when exactly one solution is correct and decays linearly as the number of correct predictions increases. We additionally impose a format reward $r^f$ to mitigate structural degradation during complex generation. This enables the proposer to effectively learn to interleave reasoning with search, yielding QA pairs that are both well-formed and challenging. The proposer prompt also requires each hop and final answer to be supported by the seed or retrieved passages (see \Cref{fig:proposer-prompt}). These grounding requirements and the absence of difficulty reward for zero-pass cases reduce, but do not eliminate, incorrect or ambiguous synthetic answers; richer signals such as retrieval uncertainty or answer ambiguity could be incorporated for less-verifiable settings.

\subsection{Solver Training}
For solver training, we sample data pairs $(x, y)$ from the proposer $\pi_\theta$ and optimize $\pi_\phi$ via group relative policy optimization (GRPO)~\citep{shao2024deepseekmath}. By computing advantages from the empirical group statistics, GRPO reinforces valid trajectories and refines the model's search and reasoning capabilities without requiring a separate value function:
\begin{equation}
\begin{aligned}
    \mathcal{J}(\phi) &= \mathbb{E}_{\substack{(x, y) \sim \pi_\theta(\cdot | \mathcal{R}), \{ \hat{y}_i \}_{i=1}^n \sim \pi_{\phi_{\mathrm{old}}}(\cdot | x, \mathcal{R})}} \\
    &[ \frac{1}{n} \sum_{i=1}^n \min \left(\frac{\pi_\phi(\hat{y}_i | x, \mathcal{R})}{\pi_{\phi_{\mathrm{old}}}(\hat{y}_i | x, \mathcal{R})} A_i, \text{clip}(\frac{\pi_\phi(\hat{y}_i | x, \mathcal{R})}{\pi_{\phi_{\mathrm{old}}}(\hat{y}_i | x, \mathcal{R})}, 1-\epsilon, 1+\epsilon) A_i \right) ]  - \beta \mathbb{D}_{KL},
\label{eq:grpo}
\end{aligned}
\end{equation}
where the advantages are computed via reward standardization (i.e., $A_i = \frac{\mathbb{I}(y = \hat{y}_i) - \texttt{mean}(\{ \mathbb{I}(y = \hat{y}_i) \}_{i=1}^n)}{\texttt{std}(\{ \mathbb{I}(y = \hat{y}_i) \}_{i=1}^n) + \delta}$). The optimization is driven by an outcome-based reward that solely evaluates the correctness of final predictions against the synthesized ground truth $y$. With increasingly complex queries from the proposer, the solver is motivated to refine its search and reasoning capabilities. Such interactions create a dynamic curriculum that supports solver improvement across diverse problem domains without curated QA training data. The solver's progress, in turn, encourages the proposer to synthesize more complex cases, establishing a feedback loop that can expand both agents' capabilities.

\subsection{The Self-Evolving \ours}
In summary, we introduce \ours, a scalable and effective framework that leverages self-evolution without curated QA training data to iteratively enhance both the proposer and solver (see \Cref{fig:method}). In each iteration, the proposer $\pi_{\theta}$ synthesizes a batch of QA pairs with heterogeneous hop structures. Utilizing solver feedback, the proposer is optimized via \hrpo to produce verifiable, diverse and challenging queries. Meanwhile, the solver leverages the generated data through GRPO to refine its search and reasoning capabilities. This alternating optimization loop creates a symbiotic feedback mechanism: as the solver improves, simple queries yield diminishing rewards, encouraging the proposer to explore more complex reasoning paths to maximize its returns. Conversely, increasingly difficult questions help sustain useful training signals, allowing the solver to continue refining its reasoning skills. Both models are initialized from the same base LLM and evolve without human-written training questions or annotated answers, relying on the external search environment to drive their performance improvements.
\section{Experiments}
\label{sec:exp}

\begin{table*}[t]
\centering
\caption{Main results of \ours against few-shot/supervised methods. We mark the best performance in bold and underline the second-best results. Unlike baselines that require curated datasets or extensive demonstrations, \ours exploits self-evolution to \emph{match or even outperform supervised search agents without curated QA training data}.}
\resizebox{\linewidth}{!}{
\begin{tabular}{@{}lcccccccc@{}}
\toprule
\multirow{2}{*}{\textbf{}}    & \textbf{NQ}    & \textbf{TriviaQA} & \textbf{PopQA} & \textbf{HotpotQA} & \textbf{2WikiMQA} & \textbf{MuSiQue} & \textbf{Bamboogle} & \textbf{Average} \\ \cmidrule(l){2-9} 
                              & \multicolumn{8}{c}{Qwen2.5-3B-Instruct}                                                                                                                \\ \midrule
Prompting                     & 0.106          & 0.288             & 0.108          & 0.149             & 0.244             & 0.020            & 0.024              & 0.134            \\
IRCoT                         & 0.111          & 0.312             & 0.200          & 0.164             & 0.171             & 0.067            & 0.240              & 0.181            \\
Search-o1                     & 0.238          & 0.472             & 0.262          & 0.221             & 0.218             & 0.054            & \textbf{0.320}     & 0.255            \\
RAG                           &  \ul{0.348}    &  \ul{0.544}       &  \ul{0.387}    & 0.255             & 0.226             & 0.047            & 0.080              & 0.270            \\
SFT                           & 0.249          & 0.292             & 0.104          & 0.186             & 0.248             & 0.044            & 0.112              & 0.176            \\
R1-Instruct                   & 0.210          & 0.449             & 0.171          & 0.208             & 0.275             & 0.060            & 0.192              & 0.224            \\
Search-R1                     & 0.323          & 0.537             & 0.364          & \textbf{0.308}    & \textbf{0.336}    & \textbf{0.105}   &  \ul{0.315}        & \textbf{0.327}   \\
\textbf{\ours}                & \textbf{0.397} & \textbf{0.572}    & \textbf{0.431} &  \ul{0.298}       &  \ul{0.291}       &  \ul{0.091}      & 0.200              &  \ul{0.326}      \\ \cmidrule(l){2-9} 
\multicolumn{1}{c}{\textbf{}} & \multicolumn{8}{c}{Qwen2.5-7B-Instruct}                                                                                                                \\ \midrule
Prompting                     & 0.134          & 0.408             & 0.140          & 0.183             & 0.250             & 0.031            & 0.120              & 0.181            \\
IRCoT                         & 0.224          & 0.478             & 0.301          & 0.133             & 0.149             & 0.072            & 0.224              & 0.239            \\
Search-o1                     & 0.151          & 0.443             & 0.131          & 0.187             & 0.176             & 0.058            & 0.296              & 0.206            \\
RAG                           & 0.349          & 0.585             & 0.392          & 0.299             & 0.235             & 0.058            & 0.208              & 0.304            \\
SFT                           & 0.318          & 0.354             & 0.121          & 0.217             & 0.259             & 0.066            & 0.112              & 0.207            \\
R1-Instruct                   & 0.270          & 0.537             & 0.199          & 0.237             & 0.292             & 0.072            & 0.293              & 0.271            \\
Search-R1                     &  \ul{0.397}    &  \ul{0.606}       &  \ul{0.404}    & \textbf{0.380}    &  \ul{0.326}       & \textbf{0.168}   & \textbf{0.408}     & \textbf{0.384}   \\
\textbf{\ours}                & \textbf{0.406} & \textbf{0.608}    & \textbf{0.416} &  \ul{0.362}       & \textbf{0.347}    &  \ul{0.104}      &  \ul{0.360}        &  \ul{0.372}      \\ \bottomrule
\end{tabular}
}
\label{tab:main}
\end{table*}

\subsection{Experiment Settings}
\noindent
\textbf{Datasets \& Models.}
For evaluation, we experiment on multiple open-domain question answering benchmarks, including three one-hop datasets Natural Questions (NQ)~\citep{kwiatkowski2019natural}, TriviaQA~\citep{joshi2017triviaqa}, PopQA~\citep{mallen2022not}; and four multi-hop datasets HotpotQA~\citep{yang2018hotpotqa}, 2WikiMultihopQA (2WikiMQA)~\citep{ho2020constructing}, MuSiQue~\citep{trivedi2022musique} and Bamboogle~\citep{press2023measuring}. These datasets cover diverse search and reasoning challenges, ensuring a comprehensive evaluation of \ours across both single-turn and multi-hop scenarios. In our experiments, we use Qwen2.5 3B/7B Instruct as base LLMs for both baseline methods and \ours.

\noindent
\textbf{Baseline \& Evaluation.}
To demonstrate the efficacy of \ours, we evaluate it against several \emph{few-shot} and \emph{supervised} baseline search agents. Few-shot baselines include standard prompting, IRCoT~\citep{trivedi2023interleaving}, Search-o1~\citep{li2025search} and retrieval augmented generation (RAG)~\citep{lewis2020retrieval}. Supervised baselines consist of supervised fine-tuning (SFT), RL-based fine-tuning without search (R1)~\citep{guo2025deepseek} and the RL-based search agent Search-R1~\citep{jin2025search}. All models are evaluated using exact match with identical search engine (E5 base) and corpus settings (English Wikipedia dump). \emph{Unlike the baselines, \ours uses neither human-annotated QA training data nor demonstrations; the shared Wikipedia corpus and search engine serve as its external knowledge environment.} Further implementation details are provided in \Cref{sec:app}.

\subsection{Experiment Results}

\noindent
\textbf{Overall Performance.}
We first discuss the main evaluation results as reported in \Cref{tab:main}. Based on the presented results, we draw several key observations: 
(1)~\emph{Overall Performance}: \ours is competitive with the strongest supervised baselines across both single-hop and multi-hop benchmarks, and exceeds them on several individual tasks. These results demonstrate that \ours effectively extends autonomous self-evolution to search agents without curated QA training data.
(2)~\emph{Superiority over Few-Shot Baselines}: \ours substantially outperforms few-shot methods across nearly every benchmark. For instance, using Qwen2.5-3B on NQ, \ours achieves 0.397 EM, significantly outperforming few-shot prompting (0.106), IRCoT (0.111) and Search-o1 (0.238). Unlike such baselines, our iterative training loop dynamically refines search capabilities across all tasks.
(3)~\emph{Parity with Supervised Search Agents}: Remarkably, \ours excels \emph{without human-annotated QA training data}. On single-hop tasks (i.e., NQ, TriviaQA and PopQA), our 3B model outperforms the supervised Search-R1 by 22.9\%, 6.5\% and 18.4\% respectively. In complex multi-hop scenarios, the 7B variant achieves roughly 90\% of Search-R1's performance and even outperforms it on the challenging 2WikiMQA.
(4)~\emph{Scaling and Self-Evolution}: As the base model scales from 3B to 7B, \ours exhibits robust gains, particularly in multi-hop reasoning. By utilizing \hrpo and a difficulty-guided reward, our framework generates a high-quality curriculum that supports gains across the initial self-evolution iterations.
Overall, our findings demonstrate that \ours effectively leverages the proposer-solver dynamics to enhance agentic search and evidence-grounded answering across diverse benchmarks; our evaluation does not separately isolate these gains from reasoning over retrieved passages.

\begin{table*}[t]
\centering
\caption{Performance comparison between \ours and data-free baseline methods.}
\resizebox{0.95\linewidth}{!}{
\begin{tabular}{@{}lcccccccc@{}}
\toprule
\multirow{2}{*}{\textbf{}} & \textbf{NQ}    & \textbf{TriviaQA} & \textbf{PopQA} & \textbf{HotpotQA} & \textbf{2WikiMQA} & \textbf{MuSiQue} & \textbf{Bamboogle} & \textbf{Average} \\ \cmidrule(l){2-9} 
                           & \multicolumn{8}{c}{Qwen2.5-3B-Instruct}                                                                                                                \\ \midrule
\textbf{SQLM*}             & 0.264          & 0.432             & 0.258          & 0.226             & 0.238             & 0.060            & 0.158              & 0.233            \\
\textbf{R-Zero*}           & 0.389          & 0.513             & 0.370          & 0.243             & 0.128             & 0.052            & 0.096              & 0.256            \\
\textbf{\ours}             & \textbf{0.397} & \textbf{0.572}    & \textbf{0.431} & \textbf{0.298}    & \textbf{0.291}    & \textbf{0.091}   & \textbf{0.200}     & \textbf{0.326}   \\ \bottomrule
\end{tabular}
}
\label{tab:baseline}
\end{table*}

\noindent
\textbf{Comparison to Data-Free Baselines.}
We further compare \ours against existing data-free methods, specifically self-questioning language models (SQLM) and self-evolving reasoning LLMs (R-Zero)~\citep{chen2025self, huang2025r}. \emph{To ensure a fair comparison, we augment baselines with multi-turn reasoning and search capabilities, denoting them SQLM* and R-Zero*.} Experiment results on the 3B backbone are reported in \Cref{tab:baseline}. From these results, we observe the following:
(1)~\ours consistently performs the best across all tasks, confirming that our proposer and training pipeline are highly effective for data-free search agents. It exceeds SQLM* and R-Zero* by an average of 39.9\% and 27.3\% respectively.
(2)~Unlike the zero-variance filtering used by SQLM*, our pass-rate reward explicitly adapts to the current solver: questions for which all solver attempts are correct or incorrect receive no difficulty reward, while partially solved questions are favored. This design provides a direct signal for producing challenging yet solvable questions.
(3)~\ours achieves superior performance and higher efficiency through the proposed \hrpo and reward formulation. Specifically, hop-based clustering significantly reduces proposer training costs without compromising performance. Correspondingly, \ours yields an average relative gain of 83.3\% over R-Zero* across the four multi-hop benchmarks.
Overall, by integrating a multi-turn reasoning-search framework with difficulty-guided rewards and the proposed \hrpo, \ours demonstrates clear advantages over existing data-free methods.

\begin{figure*}[t]
    \centering
    \includegraphics[trim=0 0 0 0, clip, width=0.95\linewidth]{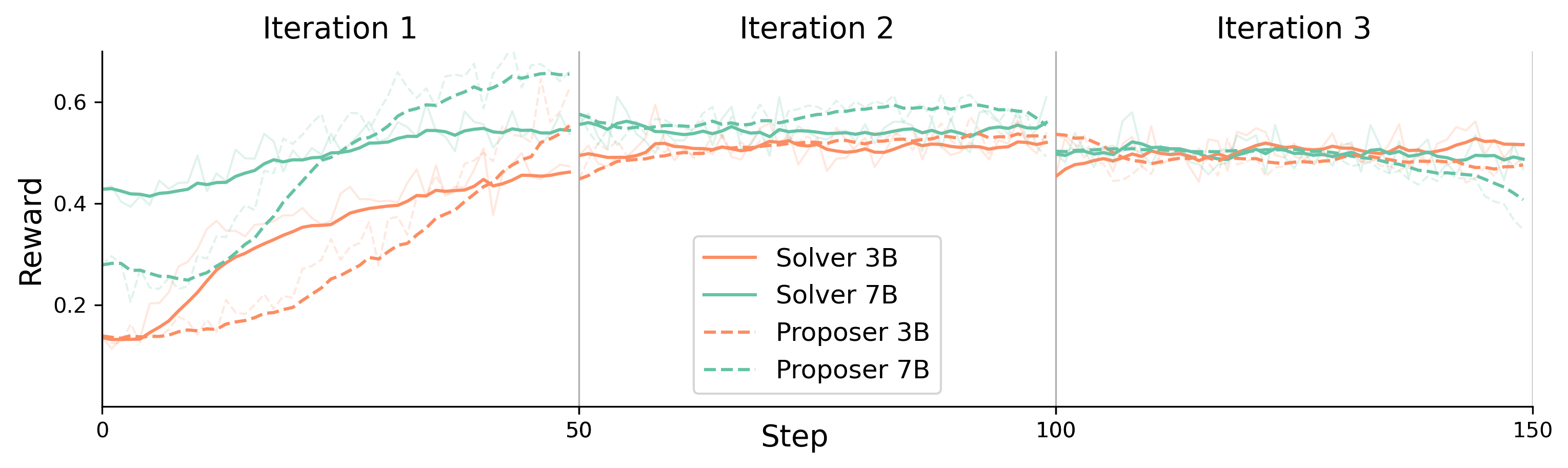}
    \caption{Iterative reward dynamics of the proposer and solver in \ours. The downward shifts in baseline rewards across iterations reflect the co-evolution of the models; as one model strengthens, it naturally lowers the initial reward floor for the other, thereby driving further self-improvement through reinforcement learning.}
    \label{fig:dynamics}
\end{figure*}

\begin{table*}[t]
\centering
\caption{Learning dynamics of \ours with increasing iterations, we mark the best performance in bold.}
\resizebox{0.95\linewidth}{!}{
\begin{tabular}{@{}lcccccccc@{}}
\toprule
\multirow{2}{*}{\textbf{}}    & \textbf{NQ}    & \textbf{TriviaQA} & \textbf{PopQA} & \textbf{HotpotQA} & \textbf{2WikiMQA} & \textbf{MuSiQue} & \textbf{Bamboogle} & \textbf{Average} \\ \cmidrule(l){2-9} 
                              & \multicolumn{8}{c}{Qwen2.5-3B-Instruct}                                                                                                                \\ \midrule
\textbf{\ours Iter 1}         & 0.381          & 0.526             & 0.392          & 0.284             & 0.243             & 0.084            & 0.216              & 0.304            \\
\textbf{\ours Iter 2}         & \textbf{0.401} & 0.563             & 0.408          & 0.289             & 0.255             & \textbf{0.102}   & \textbf{0.216}     & 0.319            \\
\textbf{\ours Iter 3}         & 0.397          & \textbf{0.572}    & \textbf{0.431} & \textbf{0.298}    & \textbf{0.291}    & 0.091            & 0.200              & \textbf{0.326}   \\ \cmidrule(l){2-9} 
\multicolumn{1}{c}{\textbf{}} & \multicolumn{8}{c}{Qwen2.5-7B-Instruct}                                                                                                                \\ \midrule
\textbf{\ours Iter 1}         & 0.392          & 0.597             & 0.395          & 0.347             & \textbf{0.361}    & \textbf{0.108}   & 0.360              & 0.366            \\
\textbf{\ours Iter 2}         & 0.406          & \textbf{0.608}    & \textbf{0.416} & \textbf{0.362}    & 0.347             & 0.104            & \textbf{0.360}     & \textbf{0.372}   \\
\textbf{\ours Iter 3}         & \textbf{0.416} & 0.608             & 0.412          & 0.352             & 0.319             & 0.107            & 0.320              & 0.360            \\ \bottomrule
\end{tabular}
}
\label{tab:dynamics}
\end{table*}

\noindent
\textbf{Training Dynamics.}
To better understand the self-evolving dynamics of \ours, we investigate the performance and rewards across iterations, with detailed reward curves and performance metrics summarized in \Cref{fig:dynamics} and \Cref{tab:dynamics}. These results provide several key observations:
(1)~We observe a steady upward trend in performance and rewards at the start of training. Using exclusively synthetic data from the proposer, both the 3B and 7B solvers rapidly reach a performance peak within approximately 50 steps, suggesting that the most significant gains are realized during this initial phase.
(2)~Following the first training phase, both the proposer and solver show substantial gains in search and reasoning. The second iteration yields consistent improvements across both solver sizes, with average gains of 4.93\% and 1.64\% across benchmarks. This confirms the continuous self-evolutionary trend of \ours.
(3)~After the second iteration, results vary by model size. The 3B model shows a modest 2.2\% increase, but the 7B model drops slightly from 0.372 to 0.360, indicating a performance plateau. Beyond this point, additional iterations for either model size yielded only marginal or no further improvements.
(4)~We identify several training failure modes, with the most common of which stemming from inconsistent token IDs across multi-turn search and reasoning steps. Interestingly, the 7B model exhibits this trend more frequently (e.g., 7B proposer iteration 3 in \Cref{fig:dynamics}), leading to increased training instability compared to the 3B variant. We also present a comparison with GRPO in \Cref{sec:app2}.
In summary, the results highlight the effectiveness of the proposer-solver interplay while identifying the technical constraints that limit indefinite self-evolution. These dynamics validate the design of our \ours framework, illustrating that autonomous self-evolution successfully replaces supervision for developing multi-turn reasoning and search capabilities.

\begin{figure*}[t]
    \centering
    \includegraphics[trim=0 0 0 0, clip, width=0.95\linewidth]{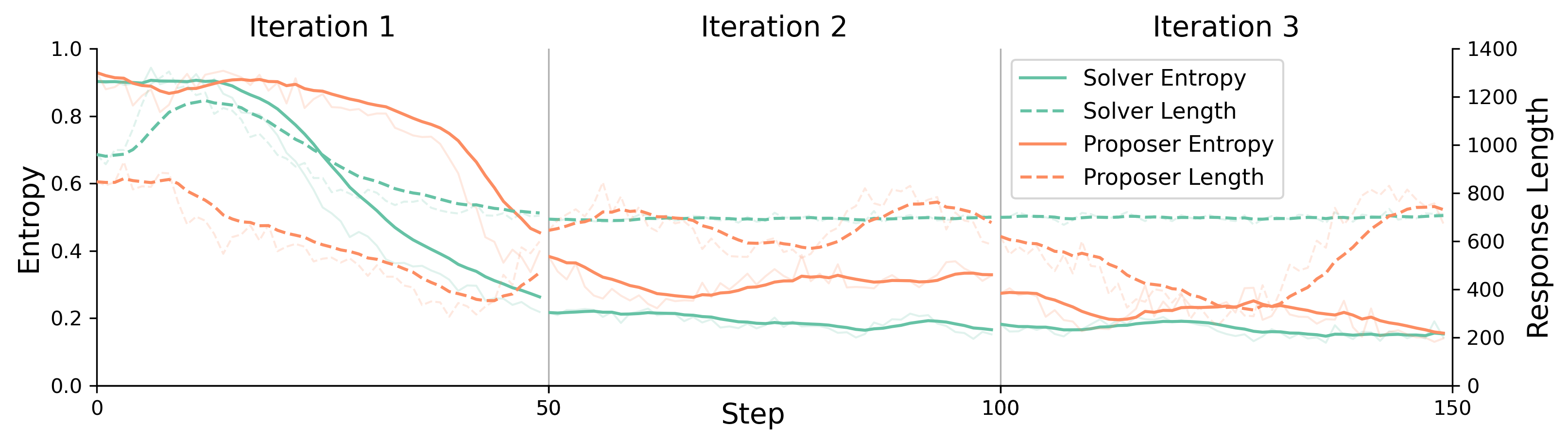}
    \caption{Averaged entropy values and response lengths of \ours 3B during training.}
    \label{fig:entropy-length}
\end{figure*}

\noindent
\textbf{Response Length and Entropy.}
Aside from performance, we investigate response entropy and length throughout the training process. Here, we focus on \ours 3B variant and visualize the average entropy values and response length in \Cref{fig:entropy-length}, leading to several noteworthy observations:
(1)~For the solver, both entropy and response length decrease gradually before stabilizing toward the end of the first iteration. This trend indicates that the solver acquires its core search and reasoning capabilities early on, followed by steady, incremental refinements in subsequent iterations.
(2)~While the proposer follows an initial trend similar to the solver, its entropy and length fluctuate in later iterations. These variations indicate that the proposer maintain generation diversity, continuously exploring different paths to present the solver with increasingly complex challenges.
(3)~Compared to trajectory lengths and the proposer entropy, the solver's entropy values are more stable; they decrease drastically and stabilize at a low level. This trend suggests that the model learns quickly and becomes more confident as its reasoning capabilities improve.
Together, these trends confirm that the proposer-solver interplay creates a dynamic learning environment: the solver achieves rapid and stable convergence, while the proposer preserves the generation diversity to ensure a broader spectrum of challenging, high-quality questions.

\begin{table*}[t]
\centering
\caption{Performance of \ours trained with different distributions of generated questions.}
\resizebox{0.95\linewidth}{!}{
\begin{tabular}{@{}lcccccccc@{}}
\toprule
\multirow{2}{*}{\textbf{}}    & \textbf{NQ}    & \textbf{TriviaQA} & \textbf{PopQA} & \textbf{HotpotQA} & \textbf{2WikiMQA} & \textbf{MuSiQue} & \textbf{Bamboogle} & \textbf{Average} \\ \cmidrule(l){2-9} 
                              & \multicolumn{8}{c}{Qwen2.5-3B-Instruct}                                                                                                                \\ \midrule
\textbf{Ratio 1:1:1:1}        & \textbf{0.402} & 0.549             & 0.394          & 0.297             & 0.287             & 0.090            & 0.176              & 0.314            \\
\textbf{Ratio 2:1:1:1}        & 0.398          & 0.560             & 0.408          & 0.287             & 0.268             & \textbf{0.096}   & 0.192              & 0.316            \\
\textbf{Ratio 4:3:2:1}        & 0.397          & \textbf{0.572}    & \textbf{0.431} & \textbf{0.298}    & \textbf{0.291}    & 0.091            & \textbf{0.200}     & \textbf{0.326}   \\ \cmidrule(l){2-9} 
\multicolumn{1}{c}{\textbf{}} & \multicolumn{8}{c}{Qwen2.5-7B-Instruct}                                                                                                                \\ \midrule
\textbf{Ratio 1:1:1:1}        & 0.392          & 0.591             & 0.396          & 0.363             & \textbf{0.352}    & \textbf{0.116}   & 0.368              & 0.368            \\
\textbf{Ratio 2:1:1:1}        & 0.398          & 0.585             & 0.396          & \textbf{0.365}    & 0.346             & 0.109            & \textbf{0.384}     & 0.369            \\
\textbf{Ratio 4:3:2:1}        & \textbf{0.406} & \textbf{0.608}    & \textbf{0.416} & 0.362             & 0.347             & 0.104            & 0.360              & \textbf{0.372}   \\ \bottomrule
\end{tabular}
}
\label{tab:hop}
\end{table*}

\noindent
\textbf{Different Hop Ratios.}
Finally, we inspect how question composition affects the overall performance of \ours. Specifically, we vary the distribution of synthetic questions from the proposer, testing ratios of 1/2/3/4-hop questions at 1:1:1:1, 2:1:1:1 and our default 4:3:2:1. These comparative results are detailed in \Cref{tab:hop}.
Interestingly, the 3B model does not exhibit performance gains as the proportion of multi-hop questions increases. The optimal multi-hop performance is achieved using the 4:3:2:1 ratio, yielding an average 0.220 EM. This suggests that for smaller base models, strengthening fundamental search capabilities (even via single-hop queries) can more effectively improve performance than focusing exclusively on complex tasks.
In contrast, the 7B variant of \ours achieves its highest overall average with the default 4:3:2:1 composition, while the other compositions remain slightly better on some complex benchmarks. This indicates that performance depends on balancing questions across hop levels rather than simply maximizing the proportion of multi-hop questions.
In summary, both model sizes benefit from a mixed curriculum, while the most effective balance varies by model scale and benchmark.

\section{Conclusion}
We introduced \ours, a self-evolution framework that enhances the reasoning and search capabilities of language agents without curated QA training data. By utilizing an iterative proposer-solver training paradigm, \ours autonomously generates diverse and increasingly challenging short-form, verifiable open-domain questions without human-written questions or annotated answers. In addition, the proposed \hrpo effectively addresses the computational bottlenecks of multi-turn tool use, enabling efficient training by clustering structurally similar queries to estimate advantages. Experimental results demonstrate that \ours is competitive with supervised search agents and surpasses them on several question answering benchmarks. These findings validate the potential of the self-evolving \ours as a powerful paradigm for developing advanced search agents in data-scarce environments. Future work will focus on extending the stability of self-evolution to overcome performance plateaus and prevent entropy collapse in larger models. Furthermore, we plan to safeguard the self-evolution process against reward hacking and bias amplification, aiming to develop robust learning frameworks that maintain the integrity and reliability of the feedback loop even in the absence of human supervision.

\clearpage
\newpage
\bibliographystyle{assets/plainnat}
\bibliography{paper, anthology}

\clearpage
\newpage
\beginappendix


\subsection{Implementation}
\label{sec:app}
In our experiments, we implement \ours as an alternating optimization of the proposer and solver, as illustrated in \Cref{fig:intro}. We start by training the proposer and utilizing the base model as the generative reward~\cite{yang2024qwen2}. Specifically, we rollout one response per prompt and extract the corresponding question and answer. These serve as inputs to the solver to compute the reward defined in \Cref{eq:reward}. Finally, this reward is employed by \hrpo to update the proposer model, as described in \Cref{eq:hrpo}. We observe that the reward saturates within approximately 50 steps due to our efficient design. Therefore, we train the proposer for 50 steps, generate QA data on the corresponding prompts, and subsequently utilize this data to train the solver for 50 steps via GRPO~\cite{shao2024deepseekmath}. In our experiments, the proposer is configured with a default generation ratio of 4:3:2:1 for 1-, 2-, 3-, and 4-hop questions respectively. Under this setting, the solver performance typically peaks after 2 to 3 self-evolution iterations, with further training yielding only marginal improvements. Therefore, we limit training to 3 iterations (150 steps per model), which is significantly fewer than baselines like R1 and Search-R1. For further hyperparameters, we conduct a minimal search over the maximum gradient norm and KL divergence coefficient to maintain training stability. The hyperparameters are held constant in our experiments, with full details reported in \Cref{tab:hrpo-hyperparameter} and \Cref{tab:grpo-hyperparameter}.

\begin{table}[h] 
\centering
\caption{Proposer (\hrpo) hyperparameter settings.}
\resizebox{0.5\textwidth}{!}{
\begin{tabular}{lccccc}
    \toprule
    Algorithm & \multicolumn{5}{c}{\texttt{\hrpo}} \\ 
    Steps & \multicolumn{5}{c}{\texttt{50}} \\
    Optimizer & \multicolumn{5}{c}{\texttt{AdamW}} \\
    Optimizer Momentum & \multicolumn{5}{c}{\texttt{$\beta_1$, $\beta_2$ = 0.9, 0.999}} \\
    Warmup Ratio & \multicolumn{5}{c}{\texttt{0.03}} \\
    Weight Decay & \multicolumn{5}{c}{\texttt{0.01}} \\
    Learning Rate & \multicolumn{5}{c}{\texttt{5e-7, 1e-6}} \\
    Max Gradient Norm & \multicolumn{5}{c}{\texttt{0.1, 1.0}} \\
    Group size in \hrpo & \multicolumn{5}{c}{\texttt{1}} \\
    Reward size in \hrpo & \multicolumn{5}{c}{\texttt{5}} \\
    KL-Div in \hrpo & \multicolumn{5}{c}{\texttt{0}} \\
    Total Train Batch Size & \multicolumn{5}{c}{\texttt{256}} \\
    LR Scheduler & \multicolumn{5}{c}{\texttt{Constant with Warmup}} \\
    Precision (WA) & \multicolumn{5}{c}{\texttt{BF16-mixed}} \\
    Max Turn in Rollout & \multicolumn{5}{c}{\texttt{5}} \\
    Max Sequence Length & \multicolumn{5}{c}{\texttt{4096}} \\
    \bottomrule
\end{tabular}
}
\label{tab:hrpo-hyperparameter}
\end{table}

\begin{table}[h] 
\centering
\caption{Solver (GRPO) hyperparameter settings.}
\resizebox{0.5\textwidth}{!}{
\begin{tabular}{lccccc}
    \toprule
    Algorithm & \multicolumn{5}{c}{\texttt{GRPO}} \\ 
    Steps & \multicolumn{5}{c}{\texttt{50}} \\
    Optimizer & \multicolumn{5}{c}{\texttt{AdamW}} \\
    Optimizer Momentum & \multicolumn{5}{c}{\texttt{$\beta_1$, $\beta_2$ = 0.9, 0.999}} \\
    Warmup Ratio & \multicolumn{5}{c}{\texttt{0.03}} \\
    Weight Decay & \multicolumn{5}{c}{\texttt{0.01}} \\
    Learning Rate & \multicolumn{5}{c}{\texttt{1e-6}} \\
    Max Gradient Norm & \multicolumn{5}{c}{\texttt{0.1, 1.0}} \\
    Group size in GRPO & \multicolumn{5}{c}{\texttt{5}} \\
    KL-Div in GRPO & \multicolumn{5}{c}{\texttt{0, 0.001}} \\
    $\epsilon$ in GRPO & \multicolumn{5}{c}{\texttt{0.2}} \\
    Total Train Batch Size & \multicolumn{5}{c}{\texttt{256}} \\
    LR Scheduler & \multicolumn{5}{c}{\texttt{Constant with Warmup}} \\
    Precision (WA) & \multicolumn{5}{c}{\texttt{BF16-mixed}} \\
    Max Turn in Rollout & \multicolumn{5}{c}{\texttt{5}} \\
    Max Sequence Length & \multicolumn{5}{c}{\texttt{3072}} \\
    \bottomrule
\end{tabular}
}
\label{tab:grpo-hyperparameter}
\end{table}

For the format reward in \hrpo, we define four requirements: (1)~adherence to the \texttt{<think>...</think>} structure; (2)~valid tool usage, including correct tool call and arguments; (3)~an extractable question enclosed in \texttt{<question>...</question>} tags; and (4)~an extractable answer within \texttt{<answer>...</answer>} tags. These components are computed individually with a lower bound of 0 and sum to a maximum total of 0.5. This stands in contrast to the difficulty score, which ranges from 0 to 1 (see \Cref{eq:reward}). We detail the tool instructions and prompts for the proposer and solver in \Cref{fig:proposer-prompt} and \Cref{fig:solver-prompt}, respectively, with qualitative examples available in \Cref{sec:qualitative}. For the external search engine, we configure it by indexing the corpus and embedding queries using the E5 model~\cite{wang2022text} following~\cite{yue2024inference, jin2025search}. During inference, we perform an approximate nearest neighbor (ANN) search to retrieve the top-3 documents. These passages are subsequently formatted and returned as the tool response. Finally, for both training and evaluation, we utilize exact match to calculate the instance-level score.

\begin{table*}[t]
\centering
\caption{Ablation results of \ours.}
\resizebox{\textwidth}{!}{
\begin{tabular}{@{}lcccccccc@{}}
\toprule
\multirow{2}{*}{\textbf{}} & \textbf{NQ}    & \textbf{TriviaQA} & \textbf{PopQA} & \textbf{HotpotQA} & \textbf{2WikiMQA} & \textbf{MuSiQue} & \textbf{Bamboogle} \\ \cmidrule(l){2-8} 
                           & \multicolumn{7}{c}{Qwen2.5-3B-Instruct}                                                                                             \\ \midrule
\ours (50 steps)           & 0.381          & 0.526             & 0.392          & 0.284             & 0.243             & 0.084            & 0.216              \\
\quad w/ 100 train steps   & 0.379          & 0.552             & 0.425          & 0.281             & 0.243             & 0.064            & 0.168              \\
\quad w/o format reward    & 0.365          & 0.501             & 0.350          & 0.287             & 0.272             & 0.056            & 0.192              \\
\quad w/ parabolic reward  & 0.388          & 0.541             & 0.408          & 0.279             & 0.246             & 0.058            & 0.144              \\
\quad w/o initial document & 0.273          & 0.461             & 0.239          & 0.239             & 0.241             & 0.099            & 0.163              \\ \bottomrule
\end{tabular}
}
\label{tab:ablation}
\end{table*}

\begin{table}[h]
\centering
\caption{Comparison between \hrpo and GRPO.}
\resizebox{0.4\textwidth}{!}{
\begin{tabular}{@{}lcc@{}}
\toprule
\multicolumn{1}{l}{} & \multicolumn{1}{c}{\quad \hrpo \quad} & \multicolumn{1}{c}{\quad GRPO \quad} \\ \cmidrule(l){2-3} 
\multicolumn{1}{l}{} & \multicolumn{2}{c}{\quad Qwen2.5-3B-Instruct \quad}  \\ \midrule
NQ                   & \textbf{0.397}            & 0.361                    \\ 
TriviaQA             & \textbf{0.572}            & 0.548                    \\ 
PopQA                & \textbf{0.431}            & 0.377                    \\ 
HotpotQA             & 0.298                     & \textbf{0.303}           \\ 
2WikiMQA             & \textbf{0.291}            & 0.279                    \\ 
MuSiQue              & 0.091                     & \textbf{0.100}           \\ 
Bamboogle            & 0.200                     & \textbf{0.272}           \\ 
Average              & \textbf{0.326}            & 0.320                    \\ \bottomrule
\end{tabular}
}
\label{tab:hrpo-grpo}
\end{table}

\begin{table}[t]
\centering
\caption{Computational cost for proposer training per iteration.}
\resizebox{0.95\textwidth}{!}{
\begin{tabular}{@{}lllll@{}}
\toprule
\textbf{Method} & \textbf{Rollouts}                 & \textbf{Wall-Clock Time} & \textbf{GPU-Hours} & \textbf{GPU Util.} \\ \midrule
\textbf{HRPO}   & 6 (1 question + 5 predictions)    & 4.16h                    & 33.28              & 21.63\%            \\
\textbf{GRPO}   & 20 (4 questions + 16 predictions) & 10.58h                   & 84.64              & 17.84\%            \\ \bottomrule
\end{tabular}
}
\label{tab:efficiency_comparison}
\end{table}

\begin{table*}[t]
\centering
\caption{Significance test of \ours against supervised methods.}
\resizebox{0.95\textwidth}{!}{
\begin{tabular}{@{}lccccccc@{}}
\toprule
\multirow{2}{*}{\textbf{}}    & \textbf{NQ}    & \textbf{TriviaQA} & \textbf{PopQA} & \textbf{HotpotQA} & \textbf{2WikiMQA} & \textbf{MuSiQue} & \textbf{Bamboogle} \\ \cmidrule(l){2-8} 
                              & \multicolumn{7}{c}{Qwen2.5-3B-Instruct}                                                                                             \\ \midrule
SFT                           & 0.249          & 0.292             & 0.104          & 0.186             & 0.248             & 0.044            & 0.112              \\
R1-Instruct                   & 0.210          & 0.449             & 0.171          & 0.208             & 0.275             & 0.060            & 0.192              \\
Search-R1                     & 0.323          & 0.537             & 0.364          & 0.308             & 0.336             & 0.105            & 0.315              \\
\textbf{\ours}                & \textbf{0.382}$_{\pm.016}$ & \textbf{0.552}$_{\pm.014}$ & \textbf{0.401}$_{\pm.030}$ & 0.303$_{\pm.004}$ & 0.291$_{\pm.014}$ & 0.101$_{\pm.016}$ & 0.229$_{\pm.033}$ \\ \cmidrule(l){2-8} 
\multicolumn{1}{c}{\textbf{}} & \multicolumn{7}{c}{Qwen2.5-7B-Instruct}                                                                                             \\ \midrule
SFT                           & 0.318          & 0.354             & 0.121          & 0.217             & 0.259             & 0.066            & 0.112              \\
R1-Instruct                   & 0.270          & 0.537             & 0.199          & 0.237             & 0.292             & 0.072            & 0.293              \\
Search-R1                     & 0.397          & 0.606             & 0.404          & 0.380             & 0.326             & 0.168            & 0.408              \\
\textbf{\ours}                & 0.400$_{\pm.008}$ & 0.599$_{\pm.011}$ & 0.405$_{\pm.011}$ & 0.362$_{\pm.002}$ & \textbf{0.351}$_{\pm.007}$ & 0.112$_{\pm.006}$ & 0.352$_{\pm.033}$ \\ \bottomrule
\end{tabular}
}
\label{tab:significance}
\end{table*}

\subsection{Additional Results}
\label{sec:app2}

We first provide a detailed quantitative breakdown of the computational efficiency of \hrpo compared to standard GRPO. As noted in the main text, standard GRPO relies on asynchronous, nested rollouts, which do not scale efficiently to larger LLMs such as our 7B model. To highlight the efficiency gains, we measured the training time for the 3B model (referenced in \Cref{tab:efficiency_comparison}) using a single 8xH100 node. \hrpo requires significantly fewer rollouts per iteration compared to standard GRPO with nested sampling. This reduction translates directly to a substantial decrease in both wall-clock time and total GPU-hours per iteration, alongside an improvement in average GPU utilization\footnote{The solver training remains identical for both methods, completing in approximately 3 hours.}. 

We provide additional results validating the design choices behind \ours and \hrpo. Specifically, we assess the effectiveness of the proposer within \ours by ablating key components, such as the format reward and the difficulty-based reward. To accommodate computational constraints, we limit the training of the 3B backbone to a single iteration while selectively removing components from the full framework. We examine the following variations: (1)~extending training to 100 steps; (2)~removing the format reward; (3)~employing a parabolic reward in the proposer (which peaks when solver accuracy is approximately 50\%); and (4)~omitting the initial document. The ablation results are reported in \Cref{tab:ablation}, based on which we observe:
(1)~increasing the training duration from 50 to 100 steps results in comparable or slightly degraded average performance (0.304 vs. 0.302), suggesting that \ours converges efficiently and does not benefit from prolonged training in each self-evolution iteration.
(2)~Removing the format reward results in an average performance drop from 0.304 to 0.289, validating its critical role in guiding the proposer to conduct effective search and reasoning while maintaining structural validity.
(3)~The default difficulty-based reward outperforms the parabolic design. This indicates that our proposed reward shaping provides more effective signals for generating challenging queries compared to a simple parabolic objective.
(4)~The most significant performance drop occurs when the initial document is removed (average score falls to 0.245). This indicates that the initial context is crucial for the model to generate diverse synthetic data.
Overall, the results demonstrate that each component of \ours plays a critical role in guiding the proposer to generate structurally valid and challenging queries, validating our integrated design as an effective framework for self-evolution without curated QA training data.

Furthermore, we compare the efficiency of \hrpo against the standard GRPO algorithm. For GRPO, we generate 4 questions per prompt and perform 4 rollouts per question, totaling 16 solver rollouts per prompt. In contrast, \hrpo generates a single question per prompt with 5 solver predictions, reducing the total rollout count to less than one-third of that required by GRPO. We conduct this comparison using the 3B backbone and present the corresponding results in \Cref{tab:hrpo-grpo}. The results show that \hrpo achieves an average score of 0.326, even surpassing the 0.320 baseline of GRPO. Notably, \hrpo demonstrates superior performance on one-hop datasets (e.g., NQ), while GRPO retains an advantage on multi-hop benchmarks (e.g., HotpotQA). This suggests that for complex multi-step reasoning tasks, higher computational resources are still essential to ensure accurate baseline estimation and maximize the learning signals from synthetic data. In summary, \hrpo achieves higher aggregate performance while utilizing significantly fewer computational resources than GRPO, highlighting its effectiveness for efficient self-evolution.

Finally, we provide additional significance testing of \ours against supervised baselines. In \Cref{tab:main}, we follow the \texttt{verl} implementation and adopt greedy decoding for evaluation. Here, we perform additional experiments to provide average results with standard deviations. The results are reported in \Cref{tab:significance} and we mark results in bold if the performance gains are \emph{statistically significant}. From the results we observe:
(1)~for the 3B backbone, \ours shows consistent gains on knowledge-intensive tasks like NQ. This indicates that self-evolution effectively optimizes search policies for direct retrieval, significantly outperforming the strong Search-R1 baseline.
(2)~On complex multi-hop datasets, the 7B backbone demonstrates more prominent gains, achieving a significant 7.67\% relative improvement on 2WikiMQA. This suggests that larger models are more effective for handling complex interleaved search and reasoning.
Overall, we find that our gains are consistent, demonstrating that \ours can indeed match or exceed supervised baselines.

\paragraph{Generated Question Analysis.}
\label{sec:question-analysis}
To directly examine the structural diversity of questions produced by existing data-free baselines, we sample and analyze 500 generated questions from SQLM* and R-Zero* using \texttt{gemini-3-flash-preview} to identify their hop structure. One-hop questions account for 93.4\% and 95.6\% of their respective outputs, confirming a strong bias toward structurally simple questions. In contrast, \ours explicitly controls structural coverage through its generation schedule: the default 4:3:2:1 ratio allocates 60\% of seed prompts to questions requiring two to four hops. We provide representative \ours generations at each hop level in \Cref{fig:proposer-example-1,fig:proposer-example-2,fig:proposer-example-3,fig:proposer-example-4}. This diagnostic uses diversity specifically to refer to coverage across question hop structures.

\subsection{Qualitative Examples}
\label{sec:qualitative}
\begin{figure*}[t]
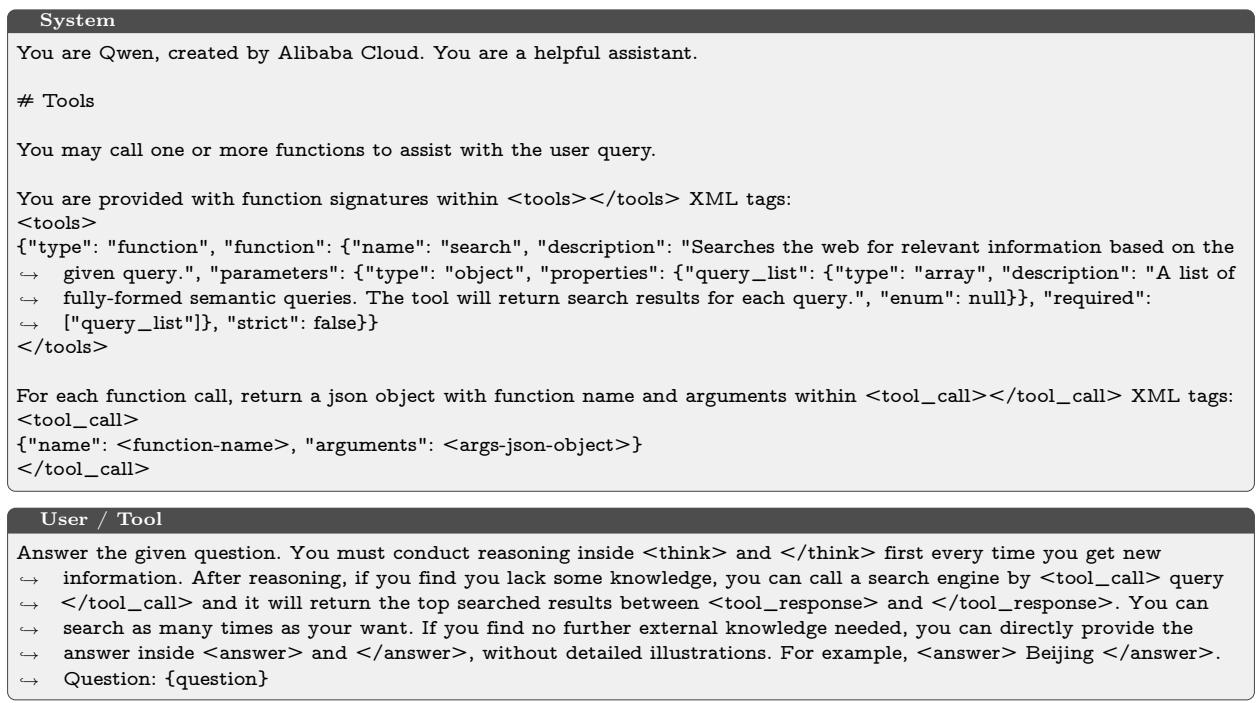

\centering

\begin{SystemBox}
\begin{ChatVerb}
You are Qwen, created by Alibaba Cloud. You are a helpful assistant.

# Tools

You may call one or more functions to assist with the user query.

You are provided with function signatures within <tools></tools> XML tags:
<tools>
{"type": "function", "function": {"name": "search", "description": "Searches the web for relevant information based on the given query.", "parameters": {"type": "object", "properties": {"query_list": {"type": "array", "description": "A list of fully-formed semantic queries. The tool will return search results for each query.", "enum": null}}, "required": ["query_list"]}, "strict": false}}
</tools>

For each function call, return a json object with function name and arguments within <tool_call></tool_call> XML tags:
<tool_call>
{"name": <function-name>, "arguments": <args-json-object>}
</tool_call>
\end{ChatVerb}
\end{SystemBox}

\begin{UserBox}
\begin{ChatVerb}
Answer the given question. You must conduct reasoning inside <think> and </think> first every time you get new information. After reasoning, if you find you lack some knowledge, you can call a search engine by <tool_call> query </tool_call> and it will return the top searched results between <tool_response> and </tool_response>. You can search as many times as your want. If you find no further external knowledge needed, you can directly provide the answer inside <answer> and </answer>, without detailed illustrations. For example, <answer> Beijing </answer>. Question: {question}
\end{ChatVerb}
\end{UserBox}

\caption{System prompt and initial instructions for the solver in \ours.}
\label{fig:solver-prompt}
\end{figure*}

\begin{figure*}[t]
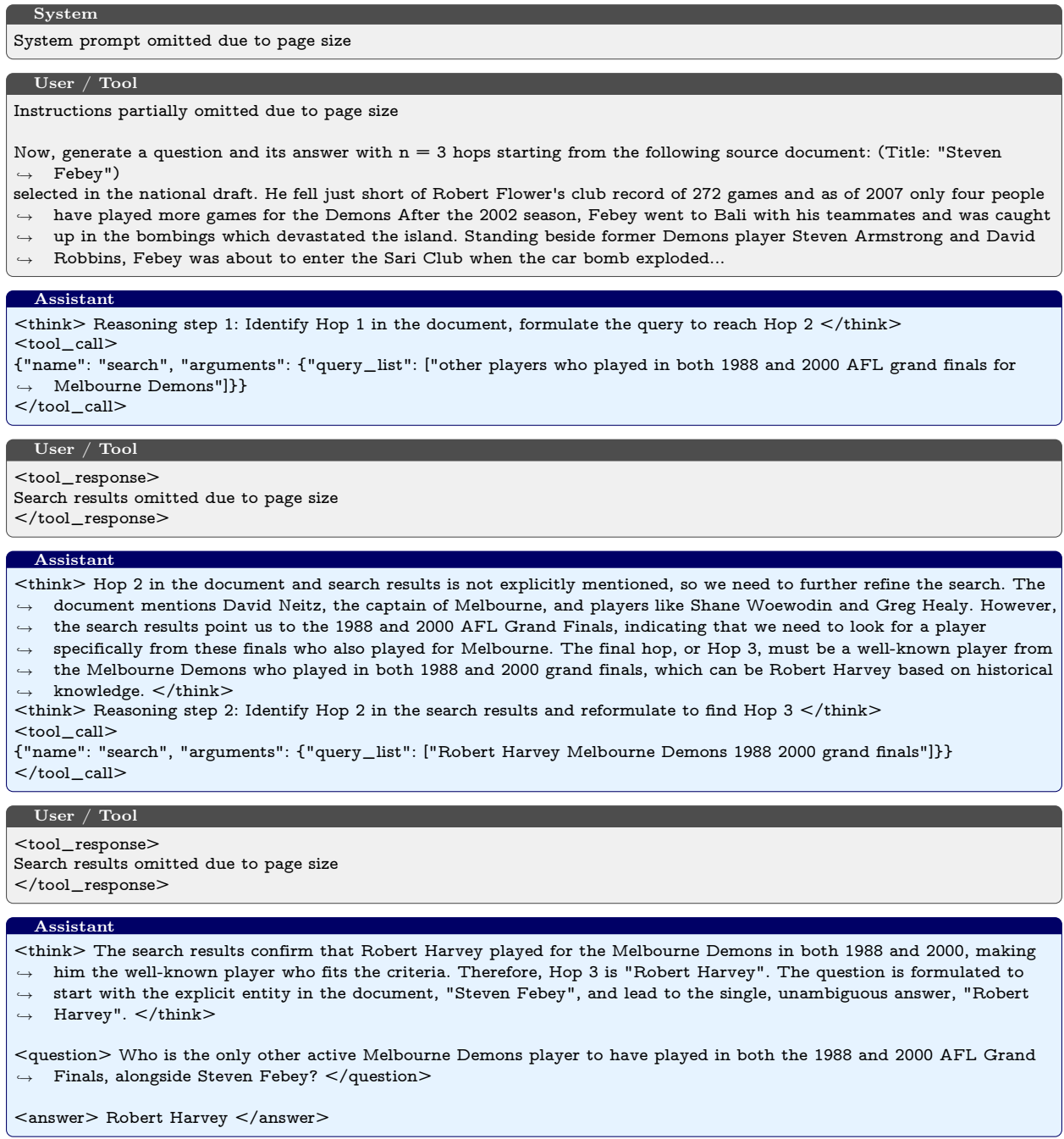

\centering

\begin{SystemBox}
\begin{ChatVerb}
System prompt omitted due to page size
\end{ChatVerb}
\end{SystemBox}

\begin{UserBox}
\begin{ChatVerb}
Instructions partially omitted due to page size

Now, generate a question and its answer with n = 3 hops starting from the following source document: (Title: "Steven Febey")
selected in the national draft. He fell just short of Robert Flower's club record of 272 games and as of 2007 only four people have played more games for the Demons After the 2002 season, Febey went to Bali with his teammates and was caught up in the bombings which devastated the island. Standing beside former Demons player Steven Armstrong and David Robbins, Febey was about to enter the Sari Club when the car bomb exploded...
\end{ChatVerb}
\end{UserBox}

\begin{AssistantBox}
\begin{ChatVerb}
<think> Reasoning step 1: Identify Hop 1 in the document, formulate the query to reach Hop 2 </think>
<tool_call>
{"name": "search", "arguments": {"query_list": ["other players who played in both 1988 and 2000 AFL grand finals for Melbourne Demons"]}}
</tool_call>
\end{ChatVerb}
\end{AssistantBox}

\begin{UserBox}
\begin{ChatVerb}
<tool_response>
Search results omitted due to page size
</tool_response>
\end{ChatVerb}
\end{UserBox}

\begin{AssistantBox}
\begin{ChatVerb}
<think> Hop 2 in the document and search results is not explicitly mentioned, so we need to further refine the search. The document mentions David Neitz, the captain of Melbourne, and players like Shane Woewodin and Greg Healy. However, the search results point us to the 1988 and 2000 AFL Grand Finals, indicating that we need to look for a player specifically from these finals who also played for Melbourne. The final hop, or Hop 3, must be a well-known player from the Melbourne Demons who played in both 1988 and 2000 grand finals, which can be Robert Harvey based on historical knowledge. </think>
<think> Reasoning step 2: Identify Hop 2 in the search results and reformulate to find Hop 3 </think>
<tool_call>
{"name": "search", "arguments": {"query_list": ["Robert Harvey Melbourne Demons 1988 2000 grand finals"]}}
</tool_call>
\end{ChatVerb}
\end{AssistantBox}

\begin{UserBox}
\begin{ChatVerb}
<tool_response>
Search results omitted due to page size
</tool_response>
\end{ChatVerb}
\end{UserBox}

\begin{AssistantBox}
\begin{ChatVerb}
<think> The search results confirm that Robert Harvey played for the Melbourne Demons in both 1988 and 2000, making him the well-known player who fits the criteria. Therefore, Hop 3 is "Robert Harvey". The question is formulated to start with the explicit entity in the document, "Steven Febey", and lead to the single, unambiguous answer, "Robert Harvey". </think>

<question> Who is the only other active Melbourne Demons player to have played in both the 1988 and 2000 AFL Grand Finals, alongside Steven Febey? </question>

<answer> Robert Harvey </answer>
\end{ChatVerb}
\end{AssistantBox}

\caption{Proposer example with \texttt{hop = 3} and initial document title: \texttt{Steven Febey}.}
\label{fig:proposer-example-3}
\end{figure*}

In this section, we present qualitative examples to illustrate the details of the \ours framework. We begin by providing the full prompts used to guide the proposer and solver in \Cref{fig:proposer-prompt} and \Cref{fig:solver-prompt}, respectively. To demonstrate the proposer's ability to synthesize training data of varying complexity, we showcase generation trajectories across different reasoning depths from \Cref{fig:proposer-example-1,fig:proposer-example-2,fig:proposer-example-3,fig:proposer-example-4}. These examples highlight the proposer's capacity to ground questions in initial documents and extend them to more challenging ones via iterative search and reasoning. Finally, \Cref{fig:solver-example-1,fig:solver-example-2,fig:solver-example-3,fig:solver-example-4} depict the solver's inference process, verifying its ability to successfully resolve these multi-step queries through structured reasoning and external tool utilization. Note that the last examples for the proposer and solver did not yield satisfactory reasoning or outputs. These instances were specifically selected to illustrate failure modes, such as instruction deviation or truncation due to length constraints. From these qualitative examples, we make the following observations:
(1)~the proposer effectively scales from single-hop extraction to complex multi-hop synthesis. It demonstrates a sophisticated ability to identify "bridge entities" that link disparate documents, transforming a simple starting point into a challenging multi-step retrieval task.
(2)~The solver consistently utilizes the thinking blocks to decompose questions into manageable sub-queries. This internal pattern allows the model to verify intermediate facts before proceeding to the next reasoning step, as seen in \Cref{fig:solver-example-3}.
(3)~Both models exhibit high proficiency in using the search tool. The solver, in particular, demonstrates `adaptive retrieval', knowing when its internal knowledge is insufficient and formulating targeted search queries to fill such gaps.
(4)~While the framework is generally robust, the final examples highlight current limitations in long-context generation. Specifically, as the number of hops increases, the models can occasionally deviate from strict formatting constraints or reach maximum token limits, resulting in incomplete trajectories. 
Overall, these qualitative examples demonstrate that while the \ours models are capable of sophisticated multi-hop reasoning, enhancing foundational model capabilities and instruction-following robustness remains essential for extending the frontier of self-evolving LLMs toward increasingly high-complexity tasks.

\begin{figure*}[t]
\centering

\begin{SystemBoxS}
\begin{ChatVerbS}
You are Qwen, created by Alibaba Cloud. You are a helpful assistant.

# Tools

You may call one or more functions to assist with the user query.

You are provided with function signatures within <tools></tools> XML tags:
<tools>
{"type": "function", "function": {"name": "search", "description": "Searches the web for relevant information based on the given query.", "parameters": {"type": "object", "properties": {"query_list": {"type": "array", "description": "A list of fully-formed semantic queries. The tool will return search results for each query.", "enum": null}}, "required": ["query_list"]}, "strict": false}}
</tools>

For each function call, return a json object with function name and arguments within <tool_call></tool_call> XML tags:
<tool_call>
{"name": <function-name>, "arguments": <args-json-object>}
</tool_call>
\end{ChatVerbS}
\end{SystemBoxS}

\begin{UserBoxS}
\begin{ChatVerbS}
You are an expert in question generation. Craft one challenging, deterministic question and its single, unambiguous answer based on the provided source document. The logical path must start from the document and require exactly n hops (i.e., n-1 searches) to reach the final answer.

### Definitions
1. Hop: A node in the reasoning chain. Hop 1 is the starting entity found in the document. Hop n is the final answer.

### Inputs
1. n: the exact number of hops in the reasoning chain (requiring n-1 searches).
2. Source document: the full source text.

### Process & Tools
1. Analyze the Document and Select the Starting Point
  - Read and analyze the source document.
  - Select a specific entity, event or detail explicitly mentioned in the text. This entity becomes Hop 1 (the initial clue).
2. Design the Chain Forwards
  - From Hop 1 to Hop 2: Identify a factual attribute or relation of Hop 1 that is NOT in the text but can be found via search. The result is Hop 2.
  - Iterate: Continue connecting the current Hop i to the next Hop i+1 using deterministic, verifiable relation found via search.
  - Stop at Hop n: Continue this process until you have exactly n hops. Hop n must be a single, canonical final answer.
3. Reasoning & Search Protocol
  - Always reason inside `<think> ... </think>` when you plan connections or receive new information.
  - For each hop transition that requires external information, issue search query using `<tool_call> ... </tool_call>`.
  - Search results will be provided between `<tool_response> ... </tool_response>` by the system.
4. Output Format
  - Emit a numbered sequence of EXACTLY n-1 search steps. For each search i (1 to n-1), produce:
    `<think> Reasoning step i: Identify Hop i in document/search results, formulate query to reach Hop i+1 </think>`
    `<tool_call> Query to search Hop i+1 </tool_call>`
    `[Wait for search results in <tool_response> from system]`
  - After completing all searches and arriving at Hop n, output the question and final answer:
    `<think> Final reasoning step: Confirm the chain is complete with Hop n and formulate the question </think>`
    `<question> A challenging question that provides Hop 1 (the initial clue) and asks for the final answer (Hop n) </question>`
    `<answer> The single, concise final answer (Hop n) </answer>`

### Examples
1. Example template for Hop n = 1, i.e. no search:
  `<think> [Explain how Hop 1 is selected from the source document and how the question is formulated] </think>`
  `<question> [Question based solely on the text entity Hop 1] </question>`
  `<answer> [Answer (Hop 1)] </answer>`
2. Example template for Hop n = 3, i.e. 2 searches:
  `<think> [Reasoning step 1: Find Hop 1 in the source document, formulate the query to reach Hop 2] </think>`
  `<tool_call> [Search query to find Hop 2 based on Hop 1] </tool_call>`
  `[Wait for search results in <tool_response> from system]`
  `<think> [Reasoning step 2: Reason on search results to identify Hop 2 and write the next query to find Hop 3] </think>`
  `<tool_call> [Search query to find Hop 3 based on Hop 2] </tool_call>`
  `[Wait for search results in <tool_response> from system]`
  `<think> [Final reasoning step: Confirm Hop 3 in search results and formulate the question starting from Hop 1] </think>`
  `<question> [Question starting with Hop 1, requiring the solver to find Hop 2 to eventually reach the Answer (Hop 3)] </question>`
  `<answer> [Answer (Hop 3)] </answer>`

### Critical Rules
1. Start in Document: Hop 1 must be explicitly present in the source text. Every subsequent hop must be supported by the corresponding search results.
2. Search is mandatory for n > 1: Each link between hops beyond Hop 1 must use the search engine.
3. Exact search count: Emit exactly (n-1) `<tool_call>` entries, no more, no fewer.
4. No spoilers: The question must mention only Hop 1; do not include or hint at intermediate hops.
5. Clarity: The question is self-contained; the answer is concise and direct (no extra commentary, formatting or explanation).
6. Chain integrity: Each hop must depend strictly on the previous hop. No hop should be skippable or derivable without its immediate predecessor.

Now, generate a question and its answer with n = {hop} hops starting from the following source document: {document}
\end{ChatVerbS}
\end{UserBoxS}

\caption{System prompt and initial instructions for the proposer in \ours.}
\label{fig:proposer-prompt}
\end{figure*}

\begin{figure*}[t]
\centering

\begin{SystemBox}
\begin{ChatVerb}
System prompt omitted due to page size
\end{ChatVerb}
\end{SystemBox}

\begin{UserBox}
\begin{ChatVerb}
Instructions partially omitted due to page size

Now, generate a question and its answer with n = 1 hops starting from the following source document: (Title: "Robert Holmes à Court")
selling his schoolmates photographs he had taken of them, and by driving them home in exchange for their travel allowances. Holmes à Court studied forestry at the University of Auckland and Massey University, where he graduated with a Bachelor of Agricultural Science in 1962...
\end{ChatVerb}
\end{UserBox}

\begin{AssistantBox}
\begin{ChatVerb}
<think> The document mentions that Robert Holmes à Court studied forestry at the University of Auckland and Massey University. Hop 1 is Robert Holmes à Court. From this, we can deduce that the next step would be to identify which university he graduated from among these two. </think>

<question> At which university did Robert Holmes à Court graduate with a Bachelor of Agricultural Science in 1962? </question>

<answer> Massey University </answer>
\end{ChatVerb}
\end{AssistantBox}

\caption{Proposer example with \texttt{hop = 1} and initial document title: \texttt{Robert Holmes à Court}.}
\label{fig:proposer-example-1}
\end{figure*}

\begin{figure*}[t]
\centering

\begin{SystemBox}
\begin{ChatVerb}
System prompt omitted due to page size
\end{ChatVerb}
\end{SystemBox}

\begin{UserBox}
\begin{ChatVerb}
Instructions partially omitted due to page size

Now, generate a question and its answer with n = 2 hops starting from the following source document: (Title: "Equivalent carbon content")
Society adopted the critical metal parameter (Pcm) for weld cracking, which was based on the work from Ito and Bessyo, is: If some of the values are not available, the following formula is sometimes used: The carbon equivalent is a measure of the tendency of the weld to form martensite on cooling and to suffer brittle fracture. When the carbon equivalent is between 0.40 and 0.60 weld preheat may be necessary...
\end{ChatVerb}
\end{UserBox}

\begin{AssistantBox}
\begin{ChatVerb}
<think> The source document mentions the "carbon equivalent" as a measure of the tendency of welds to form martensite and suffer brittle fracture. Hop 1 is "carbon equivalent". For Hop 2, the formula to determine if a spot weld will fail needs to be found, which logically follows from the concept of carbon equivalent.</think>
<tool_call>
{"name": "search", "arguments": {"query_list": ["carbon equivalent formula to determine spot weld failure"]}}
</tool_call>
\end{ChatVerb}
\end{AssistantBox}

\begin{UserBox}
\begin{ChatVerb}
<tool_response>
Search results omitted due to page size
</tool_response>
\end{ChatVerb}
\end{UserBox}

\begin{AssistantBox}
\begin{ChatVerb}
<think> Reviewing the search results, Doc 1 directly provides the formula used to determine if a spot weld will fail, which is referenced in the source document. The formula is: The carbon equivalent formula is used to determine if a spot weld will fail. Hop 1 is "carbon equivalent", and Hop 2, which is the final answer, is the formula used to determine if a spot weld will fail.</think>

<question> What specific formula is used to determine if a spot weld will fail based on the carbon equivalent? </question>

<answer> The carbon equivalent formula </answer>
\end{ChatVerb}
\end{AssistantBox}

\caption{Proposer example with \texttt{hop = 2} and initial document title: \texttt{Equivalent carbon content}.}
\label{fig:proposer-example-2}
\end{figure*}

\begin{figure*}[t]
\centering

\begin{SystemBox}
\begin{ChatVerb}
System prompt omitted due to page size
\end{ChatVerb}
\end{SystemBox}

\begin{UserBox}
\begin{ChatVerb}
Instructions partially omitted due to page size

Now, generate a question and its answer with n = 4 hops starting from the following source document: (Title: "Australian Labor Party (Australian Capital Territory Branch)")
within the ALP. The issue was Vietnam and it involved a North-South split. The contender Gordon Walsh won the preselection, but this was later overturned by the NSW Branch...
\end{ChatVerb}
\end{UserBox}

\begin{AssistantBox}
\begin{ChatVerb}
<think> Reasoning step 1: Identify Hop 1 in the document, which is the 'Canberra City (North) Branch'. It mentions 118 members in 1968 but does not directly relate to the ACT Branch establishment. Formulate a query to find information about the ACT Branch's autonomy. </think>
<tool_call>
{"name": "search", "arguments": {"query_list": ["when did the alp national conference establish the self-governing act branch"]}}
</tool_call>
\end{ChatVerb}
\end{AssistantBox}

\begin{UserBox}
\begin{ChatVerb}
<tool_response>
Search results omitted due to page size
</tool_response>
\end{ChatVerb}
\end{UserBox}

\begin{AssistantBox}
\begin{ChatVerb}
<think> Reasoning step 2: From the search result in Document 1, it is mentioned that "ALP National Conference established an autonomous ACT Branch in 1973". This provides us with our Hop 2, which is 1973. Now, I need to use the information from Hop 2 (1973) to formulate a query for Hop 3. Since the information moves from the establishment of the ACT Branch to the self-governance of the ACT, a query about when the ACT became a self-governing territory would be appropriate. </think>
<tool_call>
{"name": "search", "arguments": {"query_list": ["when did the australian capital territory become a self-governing territory"]}}
</tool_call>
\end{ChatVerb}
\end{AssistantBox}

\begin{UserBox}
\begin{ChatVerb}
<tool_response>
Search results omitted due to page size
</tool_response>
\end{ChatVerb}
\end{UserBox}

\begin{AssistantBox}
\begin{ChatVerb}
<think> Reasoning step 3: From the search results, it is clear that the ACT became a self-governing territory in 1989. This gives us Hop 3, which is 1989. The final step is to create a query that determines the final answer (Hop 4) based on these previous hops. </think>
<tool_call>
{"name": "search", "arguments": {"query_list": ["when did the australian capital territory Legislative Assembly first sit"]}}
</tool_call>
\end{ChatVerb}
\end{AssistantBox}

\begin{UserBox}
\begin{ChatVerb}
<tool_response>
Search results omitted due to page size
</tool_response>
\end{ChatVerb}
\end{UserBox}

\begin{AssistantBox}
\begin{ChatVerb}
<think> Reasoning step 4: The search results clearly indicate that the Australian Capital Territory Legislative Assembly first sat in 1989. This is our final answer, Hop 4, which directly follows from the previous hops. The challenge in the question is designed such that the initial entity 'Canberra City (North) Branch' is provided as the starting point, and the solver needs to follow the hops as established to reach the final answer.

Now I will formulate the question and the final answer:
</think>
<question> In which year did the Australian Labor Party (ACT Branch) first sit as the Legislative Assembly of the Australian Capital Territory? </question>
<answer> 1989 </answer>
\end{ChatVerb}
\end{AssistantBox}

\caption{Proposer example with \texttt{hop = 4} and initial document title: \texttt{Australian Labor Party (Australian Capital Territory Branch)}.}
\label{fig:proposer-example-4}
\end{figure*}

\begin{figure*}[t]
\centering

\begin{SystemBox}
\begin{ChatVerb}
System prompt omitted due to page size
\end{ChatVerb}
\end{SystemBox}

\begin{UserBox}
\begin{ChatVerb}
Answer the given question. You must conduct reasoning inside <think> and </think> first every time you get new information. After reasoning, if you find you lack some knowledge, you can call a search engine by <tool_call> query </tool_call> and it will return the top searched results between <tool_response> and </tool_response>. You can search as many times as your want. If you find no further external knowledge needed, you can directly provide the answer inside <answer> and </answer>, without detailed illustrations. For example, <answer> Beijing </answer>. Question: Where was the place of burial of Charles Mathew's father?
\end{ChatVerb}
\end{UserBox}

\begin{AssistantBox}
\begin{ChatVerb}
I need to conduct a search to find out where Charles Mathew's father was buried.
<tool_call>
{"name": "search", "arguments": {"query_list": ["Where was the place of burial of Charles Mathew's father"]}}
</tool_call>
\end{ChatVerb}
\end{AssistantBox}

\begin{UserBox}
\begin{ChatVerb}
<tool_response>
Search results omitted due to page size
</tool_response>
\end{ChatVerb}
\end{UserBox}

\begin{AssistantBox}
\begin{ChatVerb}
According to the search results, Father Mathew, Charles Mathew's father, was buried in Cork city in a cemetery which he had himself established. 

<answer>Cork</answer>
\end{ChatVerb}
\end{AssistantBox}

\caption{Solver example for the question: \texttt{Where was the place of burial of Charles Mathew's father?}}
\label{fig:solver-example-1}
\end{figure*}

\begin{figure*}[t]
\centering

\begin{SystemBox}
\begin{ChatVerb}
System prompt omitted due to page size
\end{ChatVerb}
\end{SystemBox}

\begin{UserBox}
\begin{ChatVerb}
Answer the given question. You must conduct reasoning inside <think> and </think> first every time you get new information. After reasoning, if you find you lack some knowledge, you can call a search engine by <tool_call> query </tool_call> and it will return the top searched results between <tool_response> and </tool_response>. You can search as many times as your want. If you find no further external knowledge needed, you can directly provide the answer inside <answer> and </answer>, without detailed illustrations. For example, <answer> Beijing </answer>. Question: Where was the husband of Octavie Coudreau born?
\end{ChatVerb}
\end{UserBox}

\begin{AssistantBox}
\begin{ChatVerb}
I need to conduct a search to find out where the husband of Octavie Coudreau was born.
<tool_call>
{"name": "search", "arguments": {"query_list": ["Where was the husband of Octavie Coudreau born"]}}
</tool_call>
\end{ChatVerb}
\end{AssistantBox}

\begin{UserBox}
\begin{ChatVerb}
<tool_response>
Search results omitted due to page size
</tool_response>
\end{ChatVerb}
\end{UserBox}

\begin{AssistantBox}
\begin{ChatVerb}
According to the search results, the husband of Octavie Coudreau was Henri Coudreau. Further searching for "Where was Henri Coudreau born" is needed.
<tool_call>
{"name": "search", "arguments": {"query_list": ["Where was Henri Coudreau born"]}}
</tool_call>
\end{ChatVerb}
\end{AssistantBox}

\begin{UserBox}
\begin{ChatVerb}
<tool_response>
Search results omitted due to page size
</tool_response>
\end{ChatVerb}
\end{UserBox}

\begin{AssistantBox}
\begin{ChatVerb}
According to the search results, Henri Coudreau was born in Sonnac. Therefore, the husband of Octavie Coudreau was born in Sonnac.
<answer>Sonnac</answer>
\end{ChatVerb}
\end{AssistantBox}

\caption{Solver example for the question: \texttt{Where was the husband of Octavie Coudreau born?}}
\label{fig:solver-example-2}
\end{figure*}

\begin{figure*}[t]
\centering

\begin{SystemBox}
\begin{ChatVerb}
System prompt omitted due to page size
\end{ChatVerb}
\end{SystemBox}

\begin{UserBox}
\begin{ChatVerb}
Answer the given question. You must conduct reasoning inside <think> and </think> first every time you get new information. After reasoning, if you find you lack some knowledge, you can call a search engine by <tool_call> query </tool_call> and it will return the top searched results between <tool_response> and </tool_response>. You can search as many times as your want. If you find no further external knowledge needed, you can directly provide the answer inside <answer> and </answer>, without detailed illustrations. For example, <answer> Beijing </answer>. Question: Which film has the director who died earlier, Gold For The Caesars or Heinz In The Moon?
\end{ChatVerb}
\end{UserBox}

\begin{AssistantBox}
\begin{ChatVerb}
I need to find out the directors of the films 'Gold For The Caesars' and 'Heinz In The Moon' to determine which one was directed by the director who died earlier.
<tool_call>
{"name": "search", "arguments": {"query_list": ["which film has the director who died earlier, Gold For The Caesars or Heinz In The Moon"]}}
</tool_call>
\end{ChatVerb}
\end{AssistantBox}

\begin{UserBox}
\begin{ChatVerb}
<tool_response>
Search results omitted due to page size
</tool_response>
\end{ChatVerb}
\end{UserBox}

\begin{AssistantBox}
\begin{ChatVerb}
From the search results, it is clear that the film 'Gold for the Caesars' was directed by Andrés De Toth in the United States and both De Toth and Sabatino Ciuffini in Italy. However, the information does not specify when Andrés De Toth died.

Let's now find out the director of 'Heinz In The Moon'.
<tool_call>
{"name": "search", "arguments": {"query_list": ["director of Heinz In The Moon"]}}
</tool_call>
\end{ChatVerb}
\end{AssistantBox}

\begin{UserBox}
\begin{ChatVerb}
<tool_response>
Search results omitted due to page size
</tool_response>
\end{ChatVerb}
\end{UserBox}

\begin{AssistantBox}
\begin{ChatVerb}
The search results indicate that the director of 'Heinz In The Moon' was Melvin Bernhardt, who died on September 12, 2015. 

Now, let's find out when Andrés De Toth died.
<tool_call>
{"name": "search", "arguments": {"query_list": ["Andrés De Toth died"]}}
</tool_call>
\end{ChatVerb}
\end{AssistantBox}

\begin{UserBox}
\begin{ChatVerb}
<tool_response>
Search results omitted due to page size
</tool_response>
\end{ChatVerb}
\end{UserBox}

\begin{AssistantBox}
\begin{ChatVerb}
The search results indicate that Andrés De Toth died on November 28, 2006. Since Melvin Bernhardt died in 2015 and Andrés De Toth died in 2006, Andrés De Toth died earlier.

Therefore, the film 'Gold for the Caesars' was directed by the director who died earlier.

<answer> Gold for the Caesars </answer>
\end{ChatVerb}
\end{AssistantBox}

\caption{Solver example for the question: \texttt{Which film has the director who died earlier, Gold For The Caesars or Heinz In The Moon?}}
\label{fig:solver-example-3}
\end{figure*}

\begin{figure*}[t]
\centering

\begin{SystemBox}
\begin{ChatVerb}
System prompt omitted due to page size
\end{ChatVerb}
\end{SystemBox}

\begin{UserBox}
\begin{ChatVerb}
Answer the given question. You must conduct reasoning inside <think> and </think> first every time you get new information. After reasoning, if you find you lack some knowledge, you can call a search engine by <tool_call> query </tool_call> and it will return the top searched results between <tool_response> and </tool_response>. You can search as many times as your want. If you find no further external knowledge needed, you can directly provide the answer inside <answer> and </answer>, without detailed illustrations. For example, <answer> Beijing </answer>. Question: When did the people who first imported slaves to New York, come to the country where Lord of the Rings was filmed?
\end{ChatVerb}
\end{UserBox}

\begin{AssistantBox}
\begin{ChatVerb}
<tool_call>
{"name": "search", "arguments": {"query_list": ["when did the people who first imported slaves to New York come to the country where Lord of the Rings was filmed"]}}
</tool_call>
\end{ChatVerb}
\end{AssistantBox}

\begin{UserBox}
\begin{ChatVerb}
<tool_response>
Search results omitted due to page size
</tool_response>
\end{ChatVerb}
\end{UserBox}

\begin{AssistantBox}
\begin{ChatVerb}
Based on the search results, the first slaves were imported to New York in 1626 by the Dutch West India Company. The last slaves were freed on July 4, 1827. 

Now, I need to find out where Lord of the Rings was filmed. Let's search for that information.
<tool_call>
{"name": "search", "arguments": {"query_list": ["where was lord of the rings filmed"]}}
</tool_call>
\end{ChatVerb}
\end{AssistantBox}

\begin{UserBox}
\begin{ChatVerb}
<tool_response>
Search results omitted due to page size
</tool_response>
\end{ChatVerb}
\end{UserBox}

\begin{AssistantBox}
\begin{ChatVerb}
The Lord of the Rings film series was primarily filmed in New Zealand. 

To answer the question, we need to find out when people came to New Zealand. Let's search for the history of immigration to New Zealand.
<tool_call>
{"name": "search", "arguments": {"query_list": ["history of immigration to New Zealand"]}}
</tool_call>
\end{ChatVerb}
\end{AssistantBox}

\begin{UserBox}
\begin{ChatVerb}
<tool_response>
Search results omitted due to page size
</tool_response>
\end{ChatVerb}
\end{UserBox}

\begin{AssistantBox}
\begin{ChatVerb}
Based on the search results, the first people to settle in New Zealand were Polynesians...
\end{ChatVerb}
\end{AssistantBox}

\caption{Solver example for the question: \texttt{When did the people who first imported slaves to New York, come to the country where Lord of the Rings was filmed?}}
\label{fig:solver-example-4}
\end{figure*}

\end{document}